\documentclass[11pt,letterpaper]{article}
\usepackage[T1]{fontenc}
\usepackage[utf8]{inputenc}
\usepackage{lmodern,microtype}
\usepackage[margin=0.93in,top=0.82in,bottom=0.84in]{geometry}
\usepackage{amsmath,amssymb,amsthm,mathtools,bm}
\usepackage{graphicx,booktabs,array,caption,placeins,float}
\usepackage{enumitem,needspace,titlesec,seqsplit}
\usepackage{xcolor}
\usepackage[authoryear,round]{natbib}
\usepackage[hidelinks]{hyperref}
\renewcommand{\arraystretch}{1.13}
\numberwithin{equation}{section}
\newcommand{\C}{\mathbb{C}}
\newcommand{\J}{J_{\!\R}}

\newcommand{\pp}{\mathbb P}

\newcommand{\R}{\mathbb{R}}
\newcommand{\D}{\mathrm{D}}
\newcommand{\Ja}{J_{\!\R}}

\newtheorem{proposition}{Proposition}

\begin{document}
\begin{center}
{\Large\bfseries Nonconformal Distortion and Local Orientation Reversal in\\[2pt]
the Fourier Circuit of a Modular-Addition Transformer\par}
\vspace{1.2em}
{\normalsize Eduardo Fuentes\textsuperscript{1,2,*} \;\textperiodcentered\; Gonzalo Riera\textsuperscript{3} \;\textperiodcentered\; H\'{e}ctor Allende-Cid\textsuperscript{4,5,*}\par}
\end{center}
\vspace{1.25em}
{\bfseries Affiliations}\par
\begin{small}
\textsuperscript{1} AI Mechanisms Lab, Chile\\
\textsuperscript{2} WriteWise, Chile\\
\textsuperscript{3} Facultad de Matem\'{a}ticas, Pontificia Universidad Cat\'{o}lica de Chile, Santiago, Chile\\
\textsuperscript{4} Fraunhofer Institute for Intelligent Analysis and Information Systems IAIS, Germany\\
\textsuperscript{5} Pontificia Universidad Cat\'{o}lica de Valpara\'{i}so, Valpara\'{i}so, Chile
\par\medskip
\textsuperscript{*} Corresponding authors: Eduardo Fuentes (ef@writewise.io) and\\
H\'{e}ctor Allende-Cid (hector.allende-cid@iais.fraunhofer.de).
\end{small}
\vspace{1.0em}
\begin{center}\textbf{Abstract}\end{center}
We study the local differential geometry of a Fourier-based modular-addition circuit in a one-layer transformer whose algorithm has previously been reverse-engineered. For each of the five key Fourier frequencies, we restrict the model's multilayer perceptron (MLP) contribution to a corresponding Fourier input plane and readout, and measure its directional Beltrami coefficient under continuous input perturbations. Whereas nondegenerate restrictions of ideal complex multiplication are conformal, the trained maps exhibit nonconformal distortion (median $|\mu|=0.304$--$0.446$), spatial organization in the Beltrami argument, and regions of local orientation reversal. The median across frequencies decreases from $0.939$ at initialization to $0.390$ after training. The key-frequency planes also show lower median distortion than most random-plane controls, although their distributions overlap and the coordinate frames differ in conditioning. Across 500 checkpoints in one training run, most of the aggregate reduction in Beltrami magnitude occurs before the principal increase in the measured Fourier-circuit formation. At the individual-input level, Beltrami magnitude shows negligible association with the correct-logit margin ($|\rho|\leq0.026$). Late-checkpoint measurements across six runs show comparable distortion for most selected frequencies. These results provide a quantitative characterization of local deformation in a frequency-projected component of an identified neural circuit, complementary to its algorithmic reconstruction.

\vspace{0.8em}
\clearpage

\section{Introduction}

Mechanistic interpretability seeks to explain neural-network behavior by identifying the computations implemented by model components. Modular addition has provided a tractable setting in which such computations can be reconstructed. Small transformers trained on this task can exhibit grokking (Power et al., 2022), with generalization emerging long after the training examples have been memorized. Nanda et al. (2023) reverse-engineered a Fourier-based algorithm learned by a one-layer transformer, showing how discrete Fourier components and trigonometric identities implement modular addition. They validated the circuit through weights, activations, and Fourier-space ablations, and used it to study training dynamics. This reconstruction specifies an algorithm implemented by the network. It does not, however, determine the local differential properties of every component involved in that algorithm, particularly when its inputs are varied continuously rather than restricted to the discrete tokens seen during training.

Recent studies have also examined the geometry of learned neural representations. Engels et al. (2025) identified circular features used for modular arithmetic in language models and tested their computational role through interventions. Gurnee et al. (2025) studied how attention heads transform low-dimensional feature manifolds during a counting task. Zhong et al. (2023) reported that changes in model hyperparameters and initialization can lead to qualitatively different algorithms for modular addition, including the Clock algorithm previously described by Nanda et al. (2023) and the Pizza algorithm. In modular addition, Moisescu-Pareja et al. (2026) found geometrically and topologically equivalent representations in models with different attention mechanisms. More generally, Brandon et al. (2026) used Riemannian pullback metrics and curvature to analyze how networks learn discrete computations on continuous input manifolds. These studies connect computation to the geometry of learned representations. A different question concerns the local distortion of the map between an identified input plane and an internal readout. Even when the relevant representation and algorithm are known, that distortion must be measured from derivatives of the map itself.

The Fourier algorithm in modular addition provides a precise reference for this question. Multiplication of complex numbers represents addition of their phases, and the ideal map $G(z,w)=czw$ is holomorphic in either variable when the other is fixed. Its nondegenerate one-variable restrictions are therefore conformal. The corresponding transformer computation is not required to equal this ideal complex map under continuous perturbations of its embedding inputs. To quantify the difference, we use the Beltrami coefficient (Ahlfors, 2006), defined from Wirtinger derivatives of a real-differentiable map. Its magnitude measures departure from conformality, and its argument determines the principal stretching axis; the sign of the Jacobian determinant indicates local orientation. Conformal methods have also been used to learn embeddings from distorted observations (Peterfreund et al., 2020), while Beltrami coefficients have been used in deep-learning methods for image registration (Chen et al., 2021). Here the object of measurement is instead an internal transformer component associated with a reconstructed Fourier circuit.

We examine the local differential geometry of the Fourier-based modular-addition circuit reconstructed by Nanda et al. (2023). For each of the five key Fourier frequencies, we restrict the transformer's MLP contribution to a corresponding input plane and Fourier readout, and measure the resulting map using directional Beltrami coefficients. This allows us to characterize departures from conformality, local stretching directions, and orientation reversal. We compare median Beltrami magnitudes between key-frequency and random control planes, examine their evolution during training in one run, evaluate late checkpoints from six training runs, and test their association with the correct-logit margin. Our analysis focuses on the geometry of an identified internal computation rather than on the geometry of representations alone.

We find nonconformal distortion at all five key frequencies, with median Beltrami magnitudes below the Gaussian bilinear reference at each frequency and an overall decrease in the median across frequencies from initialization. The argument of the Beltrami coefficient exhibits spatial organization, with similar values at neighboring points, and orientation-reversing regions differ in extent across frequencies. In the main experiment, median Beltrami magnitudes on the key-frequency planes are lower than those of most random control planes, although the distributions overlap and differ in coordinate-frame conditioning. During training, Beltrami magnitude changes early, before much of the measured circuit formation, but does not follow a monotonic trajectory. At the level of individual inputs, it has little association with the correct-logit margin. Together, these results show that recovering a circuit's algorithm does not fully characterize the local geometry of its implementation. Directional Beltrami analysis provides a complementary way to measure the deformation of an internal map whose computation has already been identified.

\section{Mathematical Framework}\label{sec:framework}

The Fourier circuit identified by Nanda et al. (2023) represents modular addition through trigonometric components of $a+b$, consistent with multiplication of unit complex numbers. We study the local geometry of the corresponding computation, rather than the accuracy of its Fourier approximation. For each key frequency, we construct a two-input complex-valued map from the trained network, differentiate it with respect to one input at a time, and measure the resulting departure from conformality. This construction is interventional: the continuous inputs used for differentiation are not ordinary tokens.

\subsection{Fourier coordinates and the internal two-input map}\label{sec:framework-map}

Let $P=113$ and let $W_E\in\R^{128\times P}$ denote the embedding matrix restricted to numeric tokens. For the five key frequencies $\mathcal{K}=\{14,35,41,42,52\}$, set $\omega_k=2\pi k/P$ and define the real Fourier coefficient vectors
\begin{equation}\label{eq:fourier-frame}
 C_k=\frac{2}{P}\sum_{a=0}^{P-1}W_E[:,a]\cos(\omega_k a),\qquad
 S_k=\frac{2}{P}\sum_{a=0}^{P-1}W_E[:,a]\sin(\omega_k a).
\end{equation}
The two vectors are linearly independent for every $k\in\mathcal K$ at the analyzed checkpoint (verified numerically). They therefore define a real plane $V_k=\operatorname{span}_{\R}\{C_k,S_k\}$, together with the complex coordinate chart
\begin{equation}\label{eq:iota}
 \iota_k:\C\longrightarrow V_k,\qquad \iota_k(x+iy)=xC_k+yS_k.
\end{equation}
Its induced complex structure satisfies $J_k C_k=S_k$ and $J_k S_k=-C_k$. The sign convention is chosen relative to the sine readout so that the ideal sum-frequency computation is represented by $zw$, rather than $\bar z\bar w$. The factor $2/P$ in Eq.~\eqref{eq:fourier-frame} is fixed by discrete Fourier reconstruction: the frequency-$k$ component of the actual embedding of token $a$ is exactly $\iota_k(e^{i\omega_k a})$. Thus the unit circle specifies the embedding's actual single-frequency amplitude, not an arbitrary radius. Conformality is measured in these declared complex coordinates, not in a coordinate-free metric on the ambient $\R^{128}$; the conditioning of the frequency frames is assessed in Appendix~C.

Write $N:\R^{128}\times\R^{128}\to\R^{512}$ for the trained network's map from the residual-stream vectors at the two numeric-token positions to the MLP activation vector at the prediction position. The positional encodings at those positions are $p_0,p_1$; the residual stream of the fixed ``='' token is held fixed inside $N$. With $W_L=W_UW_{\mathrm{out}}\in\R^{113\times512}$, let $c_k,s_k\in\R^{113}$ be the unit-normalized sequences $\cos(\omega_k t)$ and $\sin(\omega_k t)$, and define
\begin{equation}\label{eq:readout}
 u_k=W_L^\top c_k,\quad v_k=W_L^\top s_k,\qquad
 \rho_k(m)=u_k^\top m+i\,v_k^\top m.
\end{equation}
The ten sine and cosine sequences for $k\in\mathcal K$ are orthonormal on $\mathbb Z/P\mathbb Z$. In particular, Eq.~\eqref{eq:readout} defines the Fourier readout of the \emph{MLP contribution} to the logits, not the transformer's complete end-to-end logit map. The map studied below is
\begin{equation}\label{eq:G}
 \boxed{\quad G_k(z,w)=\rho_k\!\left(N\bigl(p_0+\iota_k(z),\,p_1+\iota_k(w)\bigr)\right),\qquad
 G_k:\C\times\C\longrightarrow\C .\quad}
\end{equation}
Only the numeric-token embeddings are replaced by their single-frequency components; the trained weights, positional encodings, attention computation, fixed prediction-position input, and MLP are otherwise retained. Equation~\eqref{eq:G} extends the unit-circle inputs to an open complex domain so that derivatives are defined. Points away from the circle are extrapolations under the declared linear chart, not observed training inputs.

\subsection{Directional Beltrami distortion}\label{sec:framework-beltrami}

For fixed $w_0$ or $z_0$, consider the two families of coordinate-line restrictions
\begin{equation}\label{eq:slots}
 h^{(a)}_{w_0}(z)=G_k(z,w_0),\qquad
 h^{(b)}_{z_0}(w)=G_k(z_0,w).
\end{equation}
These are restrictions to complex coordinate lines; they are \emph{not} assumed holomorphic. Given a real-differentiable map $h:\C\to\C$ with $z=x+iy$, its Wirtinger derivatives are
\begin{equation}\label{eq:wirtinger}
 \partial_z h=\tfrac12(\partial_xh-i\partial_yh),\qquad
 \partial_{\bar z}h=\tfrac12(\partial_xh+i\partial_yh).
\end{equation}
At points where the slice is $C^1$ and $\partial_z h\ne0$, define
\begin{equation}\label{eq:beltrami}
 \mu_h=\frac{\partial_{\bar z}h}{\partial_zh},\qquad
 K_h=\frac{1+|\mu_h|}{1-|\mu_h|}\quad\text{only when }|\mu_h|<1.
\end{equation}
Applying Eq.~\eqref{eq:beltrami} to the two restrictions yields $\mu_k^{(a)}(z,w)$, $\mu_k^{(b)}(z,w)$ and their respective dilatations. The principal measurements in Section~3 use the first-input coefficient $\mu_k^{(a)}$ unless stated otherwise. The excluded critical locus comprises points of nondifferentiability and points where $\partial_z h=0$; we do not assume that the actual restricted network is globally $C^1$ or prove that all such loci have measure zero.

The geometry is read directly from the real $2\times2$ Jacobian $\Ja h$. The identity
\begin{equation}\label{eq:detidentity}
 \det\Ja h=|\partial_zh|^2-|\partial_{\bar z}h|^2
 =|\partial_zh|^2(1-|\mu_h|^2)
\end{equation}
shows that $|\mu_h|<1$ corresponds to local orientation preservation, $|\mu_h|>1$ to reversal, and $|\mu_h|=1$ to zero determinant, wherever $\mu_h$ is defined. Unlike in a quasiconformal map, $|\mu_k|<1$ is \emph{not imposed} on the network. We therefore report $K$ only on the orientation-preserving locus. For a nonsingular Jacobian with singular values $\sigma_1\ge\sigma_2>0$, the orientation-independent axis ratio satisfies
\begin{equation}\label{eq:axisratio}
 \frac{\sigma_1}{\sigma_2}=\frac{1+|\mu_h|}{|1-|\mu_h||}.
\end{equation}
It agrees with $K$ when $|\mu_h|<1$ but also measures finite distortion at orientation-reversing points; it diverges for a singular Jacobian. The argument of $\mu_h$ retains directional information discarded by $|\mu_h|$: the local principal stretch axis is $\tfrac12\arg\mu_h$ modulo $\pi$, when that axis is defined. Section~3 and Appendix~B report its spatial organization separately from the magnitude.

\subsection{Coordinate dependence and exact references}\label{sec:framework-invariance}

\begin{proposition}[Coordinate behavior]\label{prop:coordinate}
Let $h$ be $C^1$ and $\partial_zh\ne0$ at the point of evaluation. Multiplying its complex-valued output by any $\alpha\in\C\setminus\{0\}$ leaves $\mu_h$ unchanged. Under an input reparametrization $g(z)=Az$ with $A\in\C\setminus\{0\}$,
\begin{equation}\label{eq:equivariance}
 \mu_{h\circ g}(z)=\frac{\bar A}{A}\,\mu_h(Az),\qquad
 |\mu_{h\circ g}(z)|=|\mu_h(Az)|.
\end{equation}
For the two-slot map the corresponding chart replacement gives $G'_k(z,w)=G_k(Az,Aw)$, with the same relation for each directional coefficient at the transported point.
\end{proposition}
Thus the readout plane is invariant under complex-linear scaling and rotation, whereas input-plane changes transport the measurement to different physical points. Unit-modulus input rotations preserve the distribution of $|\mu|$ over the full torus; radial rescalings generally do not. We therefore use the fixed Fourier normalization in Eq.~\eqref{eq:fourier-frame} throughout. The directional coefficients also obey $\mu_k^{(b)}(z,w)=\mu_k^{(a)}(w,z)$ \emph{if} $G_k(z,w)=G_k(w,z)$ exactly; approximate slot symmetry in the trained model is not treated as this exact identity. Formal proofs and qualifications are given in Appendix~A.

Two references calibrate the magnitude. For the ideal complex multiplication $G_k(z,w)=c\,zw$ with $c\ne0$, every nondegenerate coordinate slice is holomorphic and therefore $\mu_k^{(a)}=\mu_k^{(b)}=0$ and $K=1$. At the opposite reference, a random real bilinear computation has an exact, distributional baseline:
\begin{proposition}[Gaussian bilinear reference]\label{prop:gaussian}
Let $B:\R^2\times\R^2\to\R^2$ be bilinear with independent standard-Gaussian tensor coefficients. For any fixed nonzero second input, the induced $2\times2$ first-input Jacobian has independent centered Gaussian entries with a common variance. Then
\begin{equation}\label{eq:gaussbaseline}
 \operatorname{median}(|\mu_B|)=1,\qquad
 \Pr\bigl(|\mu_B|<1\bigr)=\tfrac12.
\end{equation}
\end{proposition}
These values follow because the squares of the holomorphic and antiholomorphic derivative magnitudes are independent, identically distributed multiples of $\chi_2^2$. They are exact population statements for the specified Gaussian ensemble, not universal identities for every bilinear map. The proof is given in Appendix~A. Together, ideal multiplication and this Gaussian reference make the observed distortion interpretable without conflating a global Fourier fit with local conformality.

\subsection{Experimental procedures}\label{sec:framework-eval}

We use the publicly released \texttt{full\_run\_data.pth} checkpoint series for a one-layer transformer trained on addition modulo $P=113$. The model has residual width 128, four attention heads, and MLP width 512; the training fraction is 0.30 and weight decay is 1. The main trained-network measurements use epoch 40,000. Details of the model and checkpoint verification are given in Appendix~E.

Automatic differentiation of $\rho_k\circ N$ with respect to the first residual-stream input gives $J^{(a)}_{\rho N}\in\R^{2\times128}$. The directional Jacobian follows from the chain rule:
\begin{equation}\label{eq:jacobian-eval}
 \Ja h^{(a)}_w(z)=J^{(a)}_{\rho N}
 \begin{bmatrix}C_k&S_k\end{bmatrix}\in\R^{2\times2}.
\end{equation}
The second-input derivative is analogous. We compute $\mu$ from these matrices in float64, excluding numerically degenerate derivatives by a relative threshold of $10^{-12}$. \emph{Regime A} samples the unit torus on a $16\times16$ grid ($12\times12$ for the random-plane comparison; $128\times128$ for pointwise maps); \emph{Regime B} uses actual numeric-token embeddings as basepoints while retaining the same Fourier-plane differentiation directions. Both are interventional measurements (Appendix~B).

At epoch 40,000, we compare the five key frequencies with 20 norm-matched random input planes and the non-key Fourier frequency $k=31$; conditioning and scale checks are described in Appendix~C. For replication, six independently trained runs are evaluated at epochs 40,000 and 40,100, selecting the five highest-embedding-Fourier-norm frequencies separately in each run (Appendix~D). Training dynamics are measured at all 500 saved checkpoints (epochs 0--49,900 in steps of 100) and compared with the fraction of pair-centered logit variance explained by the five key Fourier frequencies. Finally, at epoch 9,400, we test the Spearman association between $|\mu|$ and the correct-logit margin of the full model on 8,939 held-out input pairs. Definitions, cutoff rules, and statistical checks for the last two analyses are provided in Appendix~E.

\setcounter{section}{2}
\setcounter{figure}{1}
\section{Results}
We characterize the geometry of the trained transformer's Fourier computation, compare the measurements with their analytical and empirical references, and examine their evolution during training. Together, these analyses distinguish properties of the internal map from their relationship to the learned Fourier circuit.

\subsection{The trained transformer exhibits nonconformal distortion}
The Fourier multiplication algorithm identified by Nanda et al. (2023) provides a conformal reference: ideal complex multiplication has $\mu=0$ wherever the relevant directional derivative is nonzero. We measured the directional Beltrami coefficient of the corresponding internal maps at the public epoch-40,000 checkpoint, whose reconstructed forward pass solves all 12,769 modular-addition pairs with 100\% accuracy (mean cross-entropy $2.4122\times10^{-7}$). Across the five key Fourier frequencies, median $|\mu|$ ranges from 0.3036 to 0.4455 on the $16\times16$ Regime A grid (Table~\ref{tab:key}). The median ratios of the largest to smallest singular values of the directional Jacobian, $\sigma_1/\sigma_2$, range from 1.8718 to 2.6070, showing that infinitesimal circles are generally deformed into ellipses.

\begin{table}[htbp]
\centering
\caption{Directional Beltrami distortion in the five key frequencies at epoch 40,000 (Regime A, $16\times16$ grid). The singular-value ratio includes orientation-reversing points.}
\label{tab:key}
\normalsize
\setlength{\tabcolsep}{9.5pt}
\begin{tabular}{@{}rrrr@{}}
\toprule
Frequency $k$ & Median $|\mu|$ & Median $\sigma_1/\sigma_2$ & Orientation preserved\\
\midrule
14 & 0.3468 & 2.0617 & 93.8\%\\
35 & 0.3902 & 2.2684 & 92.6\%\\
41 & 0.3036 & 1.8718 & 93.8\%\\
42 & 0.4455 & 2.6070 & 78.5\%\\
52 & 0.4001 & 2.2798 & 94.9\%\\
\bottomrule
\end{tabular}
\end{table}

\begin{figure}[htbp]
\centering
\includegraphics[width=0.92\linewidth]{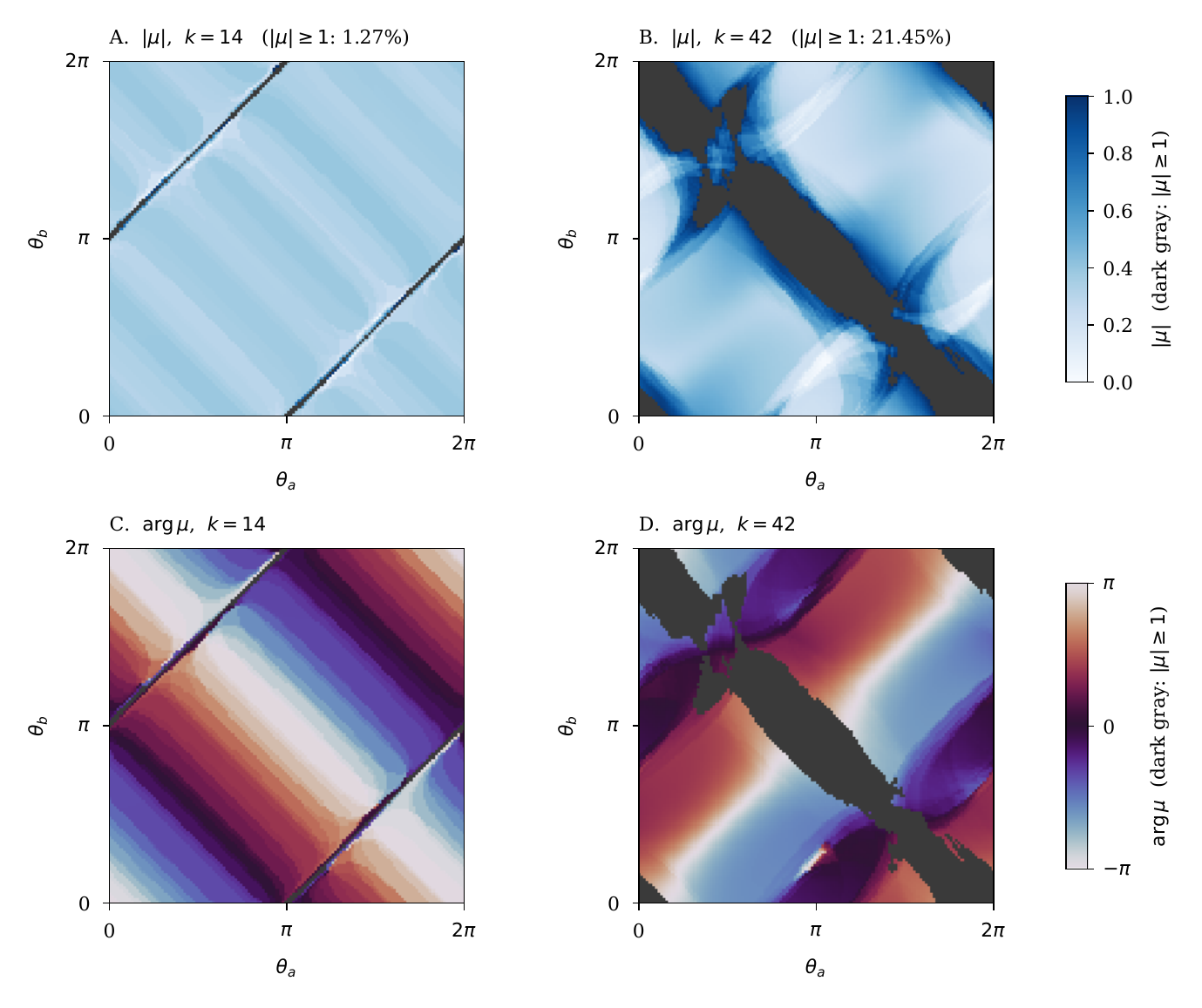}
\caption{Pointwise directional Beltrami field of the trained transformer (epoch 40,000; Regime A; $128\times128$ torus grid). (A, B) $|\mu|$ for $k=14$ and $k=42$ on the orientation-preserving locus; dark gray indicates $|\mu|\geq1$ (1.27\% and 21.45\%, respectively). (C, D) $\arg\mu$ on the same locus using a cyclic colormap. Sharp boundaries reflect the piecewise-analytic structure induced by ReLU activation regions.}
\label{fig:pointwise}
\end{figure}

The departure from conformality includes a change in local orientation. The strict condition $\|\mu\|_\infty<1$ fails on the sampled domain for every key frequency. On the finer $128\times128$ grid, the fractions of points with $|\mu|\geq1$ are 1.27\%, 8.91\%, 5.78\%, 21.45\%, and 4.08\% for $k=14,35,41,42,52$, respectively. These regions are visible in Figure~\ref{fig:pointwise}A--B: the orientation-reversing region for $k=14$ is concentrated on the antipodal-input line and its immediately neighboring grid diagonals, while $k=42$ exhibits a much larger reversing region. At $k=14$, all 208 such grid points fall within this three-diagonal band. The conventional dilatation $K$ is restricted to $|\mu|<1$; in the present measurements its censored medians (approximately 2) underestimate the median distortion of the full sampled domains, whose uncensored axis ratios are reported in Table~\ref{tab:key} (Appendix B).

The distortion also exhibits spatial organization. The axial orientation of $\mu$ is defined by $\theta=\tfrac12\arg\mu\pmod\pi$, with the angular analysis restricted to $|\mu|<1$. Although the global axial resultant lengths are small (0.0124--0.2019), neighboring grid points have angular concentrations of 0.7715--0.9800, compared with permutation-null means of 0.0412--0.0524. The observed concentrations exceed those of all 1,000 spatial permutations, consistent with the angular fields shown in Figure~\ref{fig:pointwise}C--D. The permutation comparison establishes local spatial organization, rather than global smoothness; full angular measurements and the separate equivariance check are provided in Appendix B.

To assess whether the measurements depend on the isolated Fourier-component inputs of Regime A, we evaluated the directional Beltrami coefficient around 40 actual embedding pairs (Regime B). Median $|\mu|$ values for $k=14,35,41,42,52$ are 0.3223, 0.4432, 0.3077, 0.4525, and 0.3550, respectively, broadly comparable with Regime A; orientation-preserving fractions range from 0.675 ($k=42$) to 1.000 ($k=14,41$). Thus, similar distortions are also measured at real embedding basepoints, although the infinitesimal perturbation directions remain interventional (Appendix B).

\subsection{Key Fourier frequencies show lower distortion than controls}
Having identified nonconformal, spatially organized geometry in the trained model, we next compare its magnitude with initialization, random planes, and independent training runs. The aggregate distortion is substantially lower after training. At initialization, the median across the five per-frequency medians is $|\mu|=0.9394$, and the mean orientation-preserving fraction is 0.548; at epoch 40,000, these values are 0.3902 and 0.907 (Table~\ref{tab:controls}, Block A). For reference, a real bilinear map with an i.i.d. standard-Gaussian directional Jacobian has population median $|\mu|=1$ and preserves orientation with probability $1/2$ (Appendix A). The aggregate shift is not uniform across frequencies: $k=14$ and $k=52$ begin at 0.4758 and 0.4035, respectively, relatively close to their final values.

\begin{table}[htbp]
\centering
\caption{Analytical and empirical controls. Block A uses the Regime A $16\times16$ grid; Block B uses a separate $12\times12$ plane-comparison experiment. Summary statistics have different per-plane aggregation rules; definitions and comparability checks are provided in Appendix C.}
\label{tab:controls}
\normalsize
\setlength{\tabcolsep}{6.4pt}
\begin{tabular}{@{}llrr@{}}
\toprule
Block & Condition & Median $|\mu|$ & Orientation preserved\\
\midrule
A & Gaussian bilinear reference (population) & 1.000 & 0.500\\
A & Untrained weights, epoch 0 & 0.939 & 0.548\\
A & Trained weights, epoch 40,000 & 0.390 & 0.907\\
\addlinespace[3pt]
B & Five key-frequency planes & 0.302--0.435 & 0.903\\
B & Twenty random control planes & 0.697 & 0.799\\
\bottomrule
\end{tabular}
\parbox{\linewidth}{\small\vspace{4pt}\emph{Aggregation:} In A, transformer $|\mu|$ values are medians of five frequency medians and orientation fractions are means; the Gaussian row gives population values. In B, the key $|\mu|$ value is a range and the orientation fraction a mean, whereas random-plane entries are medians of per-plane statistics (Appendix C).}
\end{table}

Within the trained model, the key Fourier planes also show lower distortion than random planes. On the separate $12\times12$ grid, the key-frequency medians are 0.302--0.435, whereas the median of medians for twenty norm-matched random planes is 0.697 (Table~\ref{tab:controls}, Block B). Three random controls (0.401, 0.430, and 0.431) overlap the key-frequency range, so the distributions are not fully separated. A rank comparison gives $U=4$ ($p=4.517\times10^{-4}$); this is a descriptive comparison because the key frequencies were functionally selected and independence of control medians within the shared network has not been established. The random controls also differ from the key planes in input-frame conditioning (median $\kappa[C\ S]$ 1.34 versus 1.00--1.03) and readout-frame conditioning. Nevertheless, rescaling the twenty random planes over $0.5$--$2.0\times s_\star$ gives median-of-medians of 0.687--0.735, while the corresponding rescaled key-frequency medians remain within 0.27--0.45; at $1.00\times s_\star$ the random-plane result is 0.697. Construction, energy-overlap checks, and scale sensitivity are detailed in Appendix C.

\begin{figure}[htbp]
\centering
\includegraphics[width=0.85\linewidth]{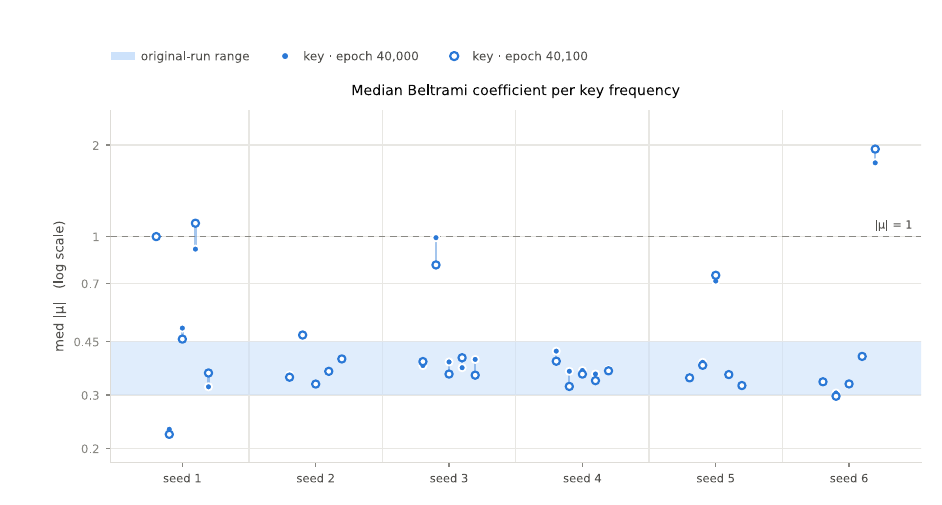}
\caption{Median $|\mu|$ for the five highest Fourier-norm frequencies of each of six training runs, measured at epochs 40,000 and 40,100. Companion dilatation and band-sensitivity panels appear in Appendix D.}
\label{fig:replicate}
\end{figure}

The geometric measurements also recur across independent training runs. For each of six runs, we selected its five highest Fourier-norm frequencies and evaluated them at epochs 40,000 and 40,100. At epoch 40,000, 22 of 30 frequency-level medians lie in $[0.30,0.45]$, while four exceed 0.90 and one is below 0.30 (Figure~\ref{fig:replicate}). Depending on the numerical band used, the replication count varies from 20 to 27 of 30. The central concentration therefore recurs across runs without every frequency converging to the same geometric range. The complete replication figure, epoch comparison, and individual results are in Appendix D.

Taken together, the initialization, control-plane, and independent-run comparisons show that lower directional distortion is a recurring feature of the key Fourier representations after training.
\FloatBarrier

\subsection{Beltrami distortion decreases before most circuit formation}
\begin{figure}[htbp]
\centering
\includegraphics[width=0.77\linewidth]{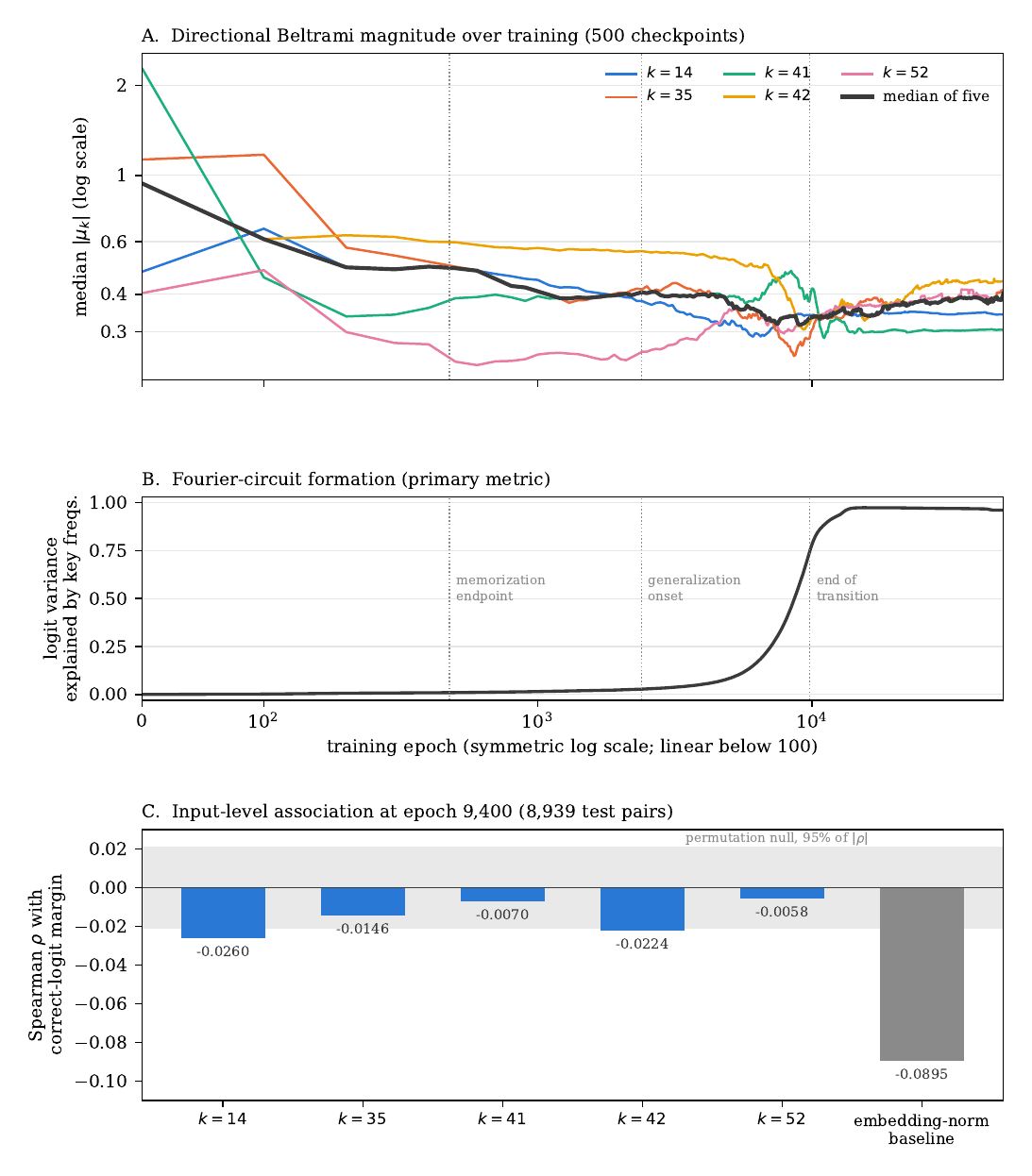}
\caption{Geometric distortion, Fourier-circuit formation, and model behavior. (A) Median $|\mu_k|$ for five key frequencies and their median at all 500 checkpoints (Regime A, $16\times16$ grid). (B) Fraction of pair-centered logit variance explained by the five key frequencies at those checkpoints. Dotted lines mark epochs 476, 2,390, and 9,830; the epoch axis is logarithmic above 100. (C) Spearman correlations of $|\mu_k|$ with correct-logit margins for 8,939 test pairs at epoch 9,400, alongside the embedding-norm baseline; shading indicates the 95th percentile of $|\rho|$ under 1,000 permutations.}
\label{fig:dynamics}
\end{figure}

To determine when this geometric change develops relative to the learned Fourier circuit, we evaluated median $|\mu_k|$ on the $16\times16$ Regime A grid at all 500 saved checkpoints, from epoch 0 to 49,900. The aggregate reduction measured between initialization and the trained checkpoint occurs early in training. Median distortion across the five frequencies is 0.9394 at initialization, 0.4920 at epoch 200, and 0.3906 at epoch 1,400, compared with 0.3902 at epoch 40,000 (Figure~\ref{fig:dynamics}A). During the main phase of Fourier-circuit formation (epochs 5,400--11,400, during which the circuit metric rises from 0.1 to 0.9), individual $|\mu_k|$ trajectories fluctuate without a common directional trend.

The circuit metric follows a different trajectory (Figure~\ref{fig:dynamics}B). At epoch 2,400, the five key frequencies explain 0.0277 of the pair-centered logit variance, compared with 0.9690 at epoch 40,000, corresponding to approximately 2.9\% of the measured development of the circuit metric. By the same epoch, 74.5--96.6\% of the measured change in $|\mu|$ has already occurred across the four frequencies with non-degenerate total change. The predefined temporal criterion required at least 60\% of total geometric change before generalization onset in at least four of five frequencies; it was met in five of five frequencies under the primary window and four of five under the sensitivity window. The exception requiring care is $k=52$: its total initial-to-final change is approximately 0.0035, making that fractional measure uninformative; excluding it yields four of four and three of four frequencies meeting the criterion, respectively. These results establish that the reduction in measured directional distortion precedes most of the development of the identified Fourier circuit (Appendix E).

We also tested whether the geometric measurement is associated with correct-logit margins on individual inputs. At epoch 9,400, using 8,939 held-out input pairs (test accuracy 0.5098; training accuracy 1.0000), the Spearman correlations for $k=14,35,41,42,52$ are $-0.0260$, $-0.0146$, $-0.0070$, $-0.0224$, and $-0.0058$, respectively (Figure~\ref{fig:dynamics}C). Two correlations, at $k=14$ and $k=42$, exceed the frequency-specific 95th percentile of $|\rho|$ under 1,000 label permutations ($\approx0.020$--$0.021$). Nevertheless, all satisfy $|\rho|\leq0.026$, far below the preregistered association threshold of 0.20. The embedding-norm baseline has $\rho=-0.0895$, and additional configurations yield a maximum $|\rho|$ of 0.052 (Appendix E). Thus, directional Beltrami magnitude does not provide a useful predictor of the correct-logit margin under the tested conditions.

\FloatBarrier

\section{Discussion}
Nanda et al. (2023) identified a Fourier-based algorithm for modular addition in a one-layer transformer by analyzing its weights, activations, and the effects of Fourier-space ablations. Here, we measure the directional Beltrami coefficient of the MLP map restricted, for each of the five key frequencies, to that frequency's input plane and Fourier readout. We find nonconformal distortion, spatially organized variation in the Beltrami field, and regions of local orientation reversal. These measurements describe the differential geometry of a component of a computation whose algorithm has already been identified.

For ideal complex multiplication, $G_k(z,w)=czw$, each nondegenerate one-variable restriction is conformal and has a zero Beltrami coefficient. The corresponding frequency-projected MLP maps of the trained transformer, however, have nonzero Beltrami coefficients and reverse orientation in parts of the sampled domain. This is consistent with Nanda et al.'s circuit reconstruction. Their analysis identifies how Fourier components and trigonometric identities implement modular addition; it does not imply that an individual projected MLP map is equal, as a differentiable function, to ideal complex multiplication under continuous embedding perturbations. The recovered algorithm and the local distortion of the projected MLP map are therefore distinct properties.

Other work has examined the geometry of representations learned by neural networks. Engels et al. (2025) identified circular features involved in modular computations in language models. Gurnee et al. (2025) showed that attention heads transform low-dimensional manifolds in a counting task. Moisescu-Pareja et al. (2026) reported that architectures with uniform and trainable attention, associated with the Clock and Pizza interpretations introduced by Zhong et al. (2023), learn geometrically and topologically equivalent representations and implement the same underlying algorithm. These studies examine the structure or functional use of representations. Our analysis instead measures derivatives of the map from Fourier input directions to an MLP readout at the same frequency. The geometry of a representation need not determine the distortion of this map.

Related geometric methods have also been applied to neural networks. Brandon et al. (2026) analyzed Riemannian pullback metrics and curvature, while Chen et al. (2021) applied Beltrami coefficients to neural image registration. We use the Beltrami coefficient to characterize the local deformation of a frequency-projected transformer component. Its magnitude is $|\partial_{\bar z}h|/|\partial_z h|$, the ratio of the magnitudes of the two Wirtinger derivatives; its argument determines a principal stretching direction, and the sign of the Jacobian determinant indicates orientation. The $\arg\mu$ fields show local spatial organization; orientation reversal is confined to a narrow antipodal band at $k=14$ but covers a much larger region at $k=42$. The five key Fourier planes also have lower median distortion than typical random-plane controls in the main experiment. Measurements from other training runs show medians in the same range as the main run for most selected frequencies at late checkpoints; the random-plane comparison was not repeated in those runs. Differences in coordinate-frame conditioning remain a possible contributor to the separation observed in the main experiment.

The training results show that directional distortion changes on a different timescale from the circuit measure. At epoch $1{,}400$, the end of the memorization phase described by Nanda et al. (2023), the median Beltrami magnitude across the five key frequencies is $0.3906$, close to its value of $0.3902$ at epoch $40{,}000$. It does not remain constant: the median and individual frequency trajectories fluctuate during subsequent training. Most of the reduction from initialization occurs before the major increase in the fraction of pair-centered logit variance explained by the key Fourier frequencies. This finding identifies an early change in the geometry of the projected MLP map. Our circuit statistic differs from the restricted and excluded losses used by Nanda et al.; the temporal ordering does not by itself imply a causal relationship between distortion and circuit formation.

The input-level analysis tests a separate relationship. At epoch $9{,}400$, Beltrami magnitude shows only weak rank correlations with the correct-logit margin across held-out input pairs. Although two correlations exceed their frequency-specific permutation thresholds, all are well below the predefined effect-size threshold. Thus, under the tested conditions, local distortion is not a useful predictor of the correct-logit margin for individual inputs. The two measurements concern different objects: the derivative of a frequency-projected MLP component and the output logits of the full model.

These findings were obtained from a one-layer transformer trained on modular addition. The temporal and input-level behavioral analyses use one training run, while measurements from six runs cover only late checkpoints. Regime B evaluates derivatives at actual embedding basepoints but along continuous Fourier directions. The random control planes differ from the key-frequency planes in input- and readout-frame conditioning. The analysis concerns two-dimensional sections of a projected MLP map rather than the full high-dimensional model. The input scale $2/P$ is fixed by the embedding reconstruction, and the transformation of the Beltrami coefficient under coordinate changes is given by Proposition 2.1.

A natural next step is to compare the directional distortion of circuits that learn equivalent representations. In particular, the uniform- and trainable-attention models examined by Moisescu-Pareja et al. (2026) provide a test of whether equivalent representation geometry also entails similar local distortion. Further experiments with conditioning-matched controls, independent training trajectories, and targeted interventions could clarify the relationship between distortion and circuit computation. The Beltrami coefficient thus adds a quantitative measure of local deformation to the analysis of this Fourier-based modular-addition circuit.

\section{Conclusion}\label{sec:conclusion}
Taken together, our results show nonconformal distortion and regions of reversed orientation in the frequency-projected MLP contribution of a transformer trained on modular addition. Directional Beltrami analysis reveals spatial variation in the magnitude and direction of this distortion. Across training, the median $|\mu|$ decreases early, before most of the measured increase in the Fourier circuit, but continues to fluctuate thereafter. In this model, $|\mu|$ does not predict the correct-logit margin for individual inputs under the tested conditions. The results indicate that the Beltrami coefficient provides a quantitative measure of the local geometry of a circuit whose Fourier algorithm has already been reconstructed. Future work should test other circuits and model architectures, examine independent training runs, and use interventions to determine whether these geometric changes affect the learned computation.

\clearpage
\newgeometry{margin=0.78in,top=0.72in,bottom=0.72in}
\begingroup
\linespread{0.99}\fontsize{10}{12}\selectfont
\setlength{\parskip}{0.23em}
\renewcommand{\small}{\fontsize{9}{11}\selectfont}
\renewcommand{\footnotesize}{\fontsize{8}{9.5}\selectfont}
\setcounter{figure}{0}
\setcounter{table}{0}
\setlength{\abovedisplayskip}{6pt}
\setlength{\belowdisplayskip}{6pt}
\appendix
\section{Mathematical Proofs}\label{app:proofs}

This appendix provides the derivations used in Section~2. The coordinate construction and Fourier normalization are established first; the local Jacobian identities then connect the Beltrami coefficient with orientation and singular values. We finally prove Proposition~2.1 (coordinate behavior) and Proposition~2.2 (the Gaussian bilinear reference). Claims involving the directional Beltrami coefficient concern points at which the relevant coordinate slice is differentiable and its holomorphic derivative is nonzero. Numerical nondegeneracy of the measured Fourier frames is an empirical check, not an exact-arithmetic theorem.

\subsection{Discrete Fourier coordinates and the choice of orientation}\label{app:fourier}

For $P=113$ and $\omega_k=2\pi k/P$, the finite character sum gives
\begin{equation}\label{eq:A-char}
 \sum_{a=0}^{P-1}e^{2\pi i ma/P}=\begin{cases}P,&m\equiv0\pmod P,\\0,&m\not\equiv0\pmod P.\end{cases}
\end{equation}
For $1\le k,\ell\le 56$, the product-to-sum identities and Eq.~\eqref{eq:A-char} imply
\begin{equation}\label{eq:A-orth}
 \sum_{a=0}^{P-1}\cos(\omega_ka)\cos(\omega_\ell a)
 =\sum_{a=0}^{P-1}\sin(\omega_ka)\sin(\omega_\ell a)
 =\frac P2\delta_{k\ell},\qquad
 \sum_{a=0}^{P-1}\cos(\omega_ka)\sin(\omega_\ell a)=0.
\end{equation}
Both sine and cosine sequences have zero mean. Consequently, the normalized vectors $\sqrt{2/P}\cos(\omega_k\cdot)$ and $\sqrt{2/P}\sin(\omega_k\cdot)$ form an orthonormal family in $\R^P$. In particular, the five-frequency readout vectors $c_k,s_k$ in Eq.~(2.3) are exactly orthonormal as functions of the output token. Their images $W_L^\top c_k,W_L^\top s_k$ need \emph{not} be orthonormal, a distinct property of the learned weights.

Let $E_a=W_E[:,a]\in\R^{128}$ and $\overline E=P^{-1}\sum_a E_a$. Discrete Fourier inversion, with the pair $k$ and $P-k$ combined into a real component, yields
\begin{equation}\label{eq:A-inversion}
 E_a=\overline E+\sum_{k=1}^{56}\bigl(C_k\cos(\omega_ka)+S_k\sin(\omega_ka)\bigr),\quad
 C_k=\frac2P\sum_aE_a\cos(\omega_ka),\quad
 S_k=\frac2P\sum_aE_a\sin(\omega_ka).
\end{equation}
Thus the frequency-$k$ component is exactly $\iota_k(e^{i\omega_ka})$, with $\iota_k(x+iy)=xC_k+yS_k$. Equation~\eqref{eq:A-inversion} fixes the factor $2/P$: replacing it by another common factor would no longer reconstruct that component on the unit circle. For each measured $k$, the chart $\iota_k$ is an isomorphism onto $V_k$ provided $C_k$ and $S_k$ are linearly independent; this condition was checked numerically on the checkpoint.

The orientation is specified \emph{relative to the chosen sine readout}. For $z=e^{i\theta_a}$ and $w=e^{i\theta_b}$,
\begin{equation}\label{eq:A-orientation}
 zw=\cos(\theta_a+\theta_b)+i\sin(\theta_a+\theta_b)
 =\bigl(\cos\theta_a\cos\theta_b-\sin\theta_a\sin\theta_b\bigr)
 +i\bigl(\sin\theta_a\cos\theta_b+\cos\theta_a\sin\theta_b\bigr).
\end{equation}
Replacing both input charts by their conjugates changes the ideal coordinate expression to $\bar z\bar w$: its real component is unchanged, but the signs of both imaginary cross terms reverse. With the output sine convention held fixed, the positive cross terms of the Fourier multiplication mechanism select the $zw$ orientation used in Section~2. Simultaneously conjugating the input charts and output readout would instead constitute a global relabeling of this orientation, equivalent to exchanging frequencies $k$ and $P-k$.

For completeness, let the ideal slice map be $G(z,w)=c zw$ with $c\in\C^*$. With $w_0\ne0$ fixed, $h^{(a)}_{w_0}(z)=cw_0z$ has $\partial_zh^{(a)}_{w_0}=cw_0\ne0$ and $\partial_{\bar z}h^{(a)}_{w_0}=0$. Thus $\mu^{(a)}=0$ and $K^{(a)}=1$; fixing $z_0\ne0$ gives the same conclusion for the second slot. At a zero fixed input, the slice is constant and its Beltrami coefficient is undefined under Eq.~(2.7). This proves the ideal reference used in Section~2 without attributing the property to the measured network.

\subsection{Real Jacobians, orientation, and principal stretching}\label{app:jacobian}

Let $h:\C\to\C$ be $C^1$ at the evaluation point and write $\alpha=\partial_z h$ and $\beta=\partial_{\bar z}h$. Its real differential acts on a complex increment $\xi$ as
\begin{equation}\label{eq:A-local}
 \D h[\xi]=\alpha\xi+\beta\bar\xi.
\end{equation}
Indeed, with $h_x=\alpha+\beta$ and $h_y=i(\alpha-\beta)$, Eq.~\eqref{eq:A-local} agrees with $h_x\,\D x+h_y\,\D y$. Identifying $\C$ with $\R^2$, the determinant of the real Jacobian is
\begin{equation}\label{eq:A-det}
 \det\J h=\operatorname{Im}(\overline{h_x}h_y)
 =\operatorname{Re}\bigl((\bar\alpha+\bar\beta)(\alpha-\beta)\bigr)
 =|\alpha|^2-|\beta|^2.
\end{equation}
If $\alpha\ne0$, then $\mu_h=\beta/\alpha$ and
\begin{equation}\label{eq:A-detmu}
 \det\J h=|\alpha|^2(1-|\mu_h|^2).
\end{equation}
This proves Eq.~(2.8) and the distinction between local orientation preservation ($|\mu|<1$), reversal ($|\mu|>1$), and a singular real Jacobian ($|\mu|=1$), wherever $\mu$ is defined. When $\alpha=0$, the Beltrami coefficient used in this paper is undefined even if the real Jacobian is nonsingular.

For $|\xi|=1$, Eq.~\eqref{eq:A-local} implies the sharp bounds
\begin{equation}\label{eq:A-sv}
 \bigl||\alpha|-|\beta|\bigr|\le|\D h[\xi]|\le|\alpha|+|\beta|.
\end{equation}
The phase of $\xi=e^{i\vartheta}$ can align $\alpha\xi$ with $\beta\bar\xi$ or oppose them, so both bounds are attained. Hence the singular values are $\sigma_1=|\alpha|+|\beta|$ and $\sigma_2=\bigl||\alpha|-|\beta|\bigr|$. At points with $\alpha\ne0$ and $\sigma_2>0$,
\begin{equation}\label{eq:A-axisratio}
 \frac{\sigma_1}{\sigma_2}=\frac{1+|\mu_h|}{|1-|\mu_h||},
\end{equation}
which is Eq.~(2.9). On $|\mu|<1$ this equals $K=(1+|\mu|)/(1-|\mu|)$; on $|\mu|>1$ it remains a finite axis ratio but $K$ is not defined under our convention. In particular, the axis ratio decreases towards $1$ as $|\mu|\to\infty$ after the singular threshold, and should not be interpreted as a monotone extension of $K$.

When $\alpha\beta\ne0$ and the singular values differ, maximal stretching occurs when $\mu_h e^{-2i\vartheta}$ is positive real, giving the input-plane axis $\vartheta=\frac12\arg\mu_h\pmod\pi$. For $\mu_h=0$ (a conformal linear differential) there is no distinguished principal axis. This establishes the directional interpretation used in Section~2 and Appendix~B.

\subsection{Proof of Proposition 2.1: coordinate behavior}\label{app:coordinate}

Let $h$ be $C^1$ and $\partial_z h\ne0$ at the evaluation point. If $\hat h(z)=\gamma h(z)$ for $\gamma\in\C^*$, linearity of the Wirtinger operators gives $\partial_z\hat h=\gamma\partial_z h$ and $\partial_{\bar z}\hat h=\gamma\partial_{\bar z}h$. Their ratio is unchanged: $\mu_{\hat h}=\mu_h$. Both $|\mu|$ and $K$ (where defined) are therefore invariant under complex-linear output reparametrization.

For $A\in\C^*$, let $g(z)=Az$, $\zeta=Az$, and suppose $h$ is $C^1$ with $\partial_\zeta h\ne0$ at $\zeta$. The Wirtinger chain rule and $\partial_zg=A$, $\partial_{\bar z}g=0$ yield
\begin{equation}\label{eq:A-chain}
 \partial_z(h\circ g)(z)=A\,(\partial_\zeta h)(Az),\qquad
 \partial_{\bar z}(h\circ g)(z)=\bar A\,(\partial_{\bar\zeta}h)(Az).
\end{equation}
Dividing gives exactly
\begin{equation}\label{eq:A-transform}
 \mu_{h\circ g}(z)=\frac{\bar A}{A}\,\mu_h(Az),\qquad
 |\mu_{h\circ g}(z)|=|\mu_h(Az)|,
\end{equation}
with $\arg\mu$ rotated by $-2\arg A$ where its argument exists. For $\iota'_k=\iota_k\circ g$ in both input slots, the definition of the measured map gives $G'_k(z,w)=G_k(Az,Aw)$. Applying Eq.~\eqref{eq:A-chain} separately to each fixed-slot restriction proves Proposition~2.1 for both directional coefficients. In particular, these equalities compare \emph{transported points}; they do not give fixed-coordinate invariance when $|A|\ne1$.

If $|A|=1$, the map $(z,w)\mapsto(Az,Aw)$ bijectively preserves the product unit torus and its normalized Haar measure. Equation~\eqref{eq:A-transform} therefore preserves the distribution of $|\mu|$ and, on its defined locus, of $K$ over the full torus. This population statement does not assert equality of finite-grid summaries under arbitrary rotations that fail to permute the sampled grid points.

Conjugating the complex-valued output reverses this orientation. Specifically, let $\widetilde h(z)=\overline{h(z)}$. The Wirtinger derivatives satisfy
\[
 \partial_z\widetilde h=\overline{\partial_{\bar z}h},\qquad
 \partial_{\bar z}\widetilde h=\overline{\partial_z h},\qquad
 \mu_{\widetilde h}=\frac{1}{\overline{\mu_h}},\qquad
 |\mu_{\widetilde h}|=\frac{1}{|\mu_h|},
\]
where $\mu_h$ is defined and nonzero. Conjugation also reverses the real-Jacobian determinant sign: $\det\J\widetilde h=-\det\J h$. In the Fourier readout, reversing the output orientation ($v_k\mapsto -v_k$, with $u_k$ unchanged) is precisely such a conjugation. Thus the reported sign of orientation, and the magnitude $|\mu|$, depend on that readout convention unless it is externally anchored.

Finally, if $G_k(z,w)=G_k(w,z)$ \emph{exactly}, then for fixed $w_0$ the functions $u\mapsto G_k(u,w_0)$ and $u\mapsto G_k(w_0,u)$ coincide. Their Wirtinger derivatives coincide wherever defined, so $\mu_k^{(a)}(z_0,w_0)=\mu_k^{(b)}(w_0,z_0)$. Approximate slot symmetry alone yields no such exact identity or quantitative stability bound.

\subsection{Proof of Proposition 2.2: Gaussian bilinear reference}\label{app:gaussian}

Consider a real bilinear map $B:\R^2\times\R^2\to\R^2$ with tensor coefficients $T_{ij\ell}\overset{\mathrm{iid}}{\sim}N(0,1)$. Fix a nonzero second input $y=(y_1,y_2)$. The first-slot restriction $x\mapsto B(x,y)$ has real Jacobian $M(y)\in\R^{2\times2}$ with entries
\begin{equation}\label{eq:A-randommatrix}
 M_{ij}(y)=\sum_{\ell=1}^2T_{ij\ell}y_\ell,
 \qquad M_{ij}(y)\overset{\mathrm{iid}}{\sim}N(0,\|y\|^2).
\end{equation}
The four entries are independent because they depend on disjoint tensor coefficients. Write $M=(m_{ij})$ and set
\begin{equation}\label{eq:A-aabcd}
 a=m_{11}+m_{22},\quad b=m_{21}-m_{12},\quad
 c=m_{11}-m_{22},\quad d=m_{21}+m_{12}.
\end{equation}
The four corresponding coefficient vectors in $\R^4$ are orthogonal and have squared length $2$. Joint Gaussianity therefore makes $a,b,c,d$ independent $N(0,2\|y\|^2)$ variables. The complex representation of the first-slot differential is
\begin{equation}\label{eq:A-alpha-beta}
 h(z)=\alpha z+\beta\bar z,\qquad
 \alpha=\frac{a+ib}{2},\quad \beta=\frac{c+id}{2}.
\end{equation}
Consequently $\alpha$ and $\beta$ are independent and identically distributed rotationally symmetric complex Gaussians, and their squared magnitudes satisfy
\begin{equation}\label{eq:A-chi2}
 |\mu_B|^2=\frac{|\beta|^2}{|\alpha|^2}
 =\frac{c^2+d^2}{a^2+b^2}\ \overset{d}{=}\ \frac{X}{Y},\qquad X,Y\overset{\mathrm{iid}}{\sim}\chi^2_2.
\end{equation}
The event $\alpha=0$ has probability zero. Since $X$ and $Y$ are independent identically distributed exponentials up to a common scale, $U=X/(X+Y)$ is uniform on $(0,1)$. Thus, for $r\ge0$,
\begin{equation}\label{eq:A-gausscdf}
 \pp(|\mu_B|\le r)=\pp\!\left(U\le\frac{r^2}{1+r^2}\right)
 =\frac{r^2}{1+r^2}.
\end{equation}
Setting $r=1$ gives $\pp(|\mu_B|<1)=1/2$ (there is no point mass at $1$). Since the distribution function is continuous and strictly increasing, its unique median is also $1$, proving Proposition~2.2. These are population results for this specified isotropic Gaussian tensor ensemble; neither conclusion is asserted for an arbitrary deterministic bilinear map.

\subsection{Regularity and the directional chain rule}\label{app:regularity}

The map $G_k=\rho_k\circ N\circ\bigl((p_0+\iota_k)\times(p_1+\iota_k)\bigr)$ contains a softmax and a ReLU MLP. On any open region with fixed strict ReLU activation signs, its operations are compositions of affine and real-analytic functions, so its restrictions and their real Jacobians are well defined. We make no global assertion that the pullbacks of all ReLU boundaries, or the zero sets of $\partial_z h$, have measure zero on the specific four-dimensional input chart or its fixed-slot slices. Equations~(2.5)--(2.12) are used only at points where the stated derivatives are defined.

Writing $F_k=\rho_k\circ N$ and $H_k=[\,C_k\ \ S_k\,]\in\R^{128\times2}$, the derivative of $z=x+iy\mapsto p_0+\iota_k(z)$ is the constant matrix $H_k$. At a differentiability point, the ordinary real chain rule therefore gives
\begin{equation}\label{eq:A-projection}
 \J h^{(a)}_w(z)=\D_1F_k\bigl(p_0+\iota_k(z),p_1+\iota_k(w)\bigr)\,H_k,
\end{equation}
with the analogous formula for the second slot. This is Eq.~(2.12). It characterizes the measurement on the trained network; floating-point thresholds used to detect numerical degeneracy are computational conventions, not additional mathematical hypotheses.

\endgroup
\clearpage
\newgeometry{margin=0.82in,top=0.74in,bottom=0.73in}
\begingroup
\linespread{1.0}\fontsize{10}{12}\selectfont
\renewcommand{\small}{\fontsize{9}{11}\selectfont}
\renewcommand{\footnotesize}{\fontsize{8}{9.5}\selectfont}
\setlength{\parskip}{5.5pt}
\setcounter{figure}{0}
\setcounter{table}{0}
\renewcommand{\thefigure}{B\arabic{figure}}
\renewcommand{\thetable}{B\arabic{table}}
\section*{Appendix B. Extended Geometric Measurements}
\label{app:B}
The measurements summarized in Section~3.1 characterize the directional Beltrami coefficient of the Fourier computation at epoch 40,000. This appendix gives the complementary statistics needed to interpret those measurements: the effect of restricting dilatation to orientation-preserving points, the spatial organization of $\arg\mu$, measurements at actual input-embedding basepoints, and the relationship between local coefficient jumps and ReLU activation-pattern changes. Unless stated otherwise, measurements use the five key frequencies $k\in\{14,35,41,42,52\}$ and the declared Regime~A embedding and readout frames.

\subsection*{B.1. Orientation reversal and the censoring of dilatation}
The conventional quasiconformal dilatation $K=(1+|\mu|)/(1-|\mu|)$ is used only where $|\mu|<1$. Consequently, summarizing $K$ over the orientation-preserving locus omits precisely those points where the directional map reverses orientation. To describe the entire sampled domain, we additionally report the ratio $\sigma_1/\sigma_2$ of the largest to smallest singular values of the real directional Jacobian. This axis ratio equals $K$ where $|\mu|<1$ and remains defined at the observed orientation-reversing points. It should not be interpreted as an orientation indicator; that information is tracked separately by the sign of the Jacobian determinant.

\begin{table}[!htbp]
\centering
\caption{Censored and uncensored median dilatation in Regime A (epoch 40,000; $16\times16$ grid). The median of $K$ uses only points with $|\mu|<1$; the median singular-value ratio uses the full grid.}
\label{tab:B1}
\vspace{0.25em}
\small
\begin{tabular}{@{}r r r r r@{}}
\toprule
$k$ & Median $K$ ($|\mu|<1$) & Median $\sigma_1/\sigma_2$ & Difference & Excluded (\%)\\
\midrule
14 & 2.0437 & 2.0617 & $+0.018$ & 6.25\\
35 & 2.0932 & 2.2684 & $+0.175$ & 7.42\\
41 & 1.8648 & 1.8718 & $+0.007$ & 6.25\\
42 & 2.1472 & 2.6070 & $+0.460$ & 21.48\\
52 & 2.1582 & 2.2798 & $+0.122$ & 5.08\\
\bottomrule
\end{tabular}
\end{table}

As shown in Table~\ref{tab:B1}, the median full-grid axis ratio is higher than the median censored $K$ at all five frequencies. The largest difference is at $k=42$ ($2.6070$ versus $2.1472$), for which $21.48\%$ of the $16\times16$ grid is excluded from the $K$ summary. These comparisons quantify the effect of censoring \emph{in the present measurements}; they do not establish that the difference must have the same sign for other maps or samples.

\begin{figure}[!htbp]
\centering
\includegraphics[width=0.97\linewidth]{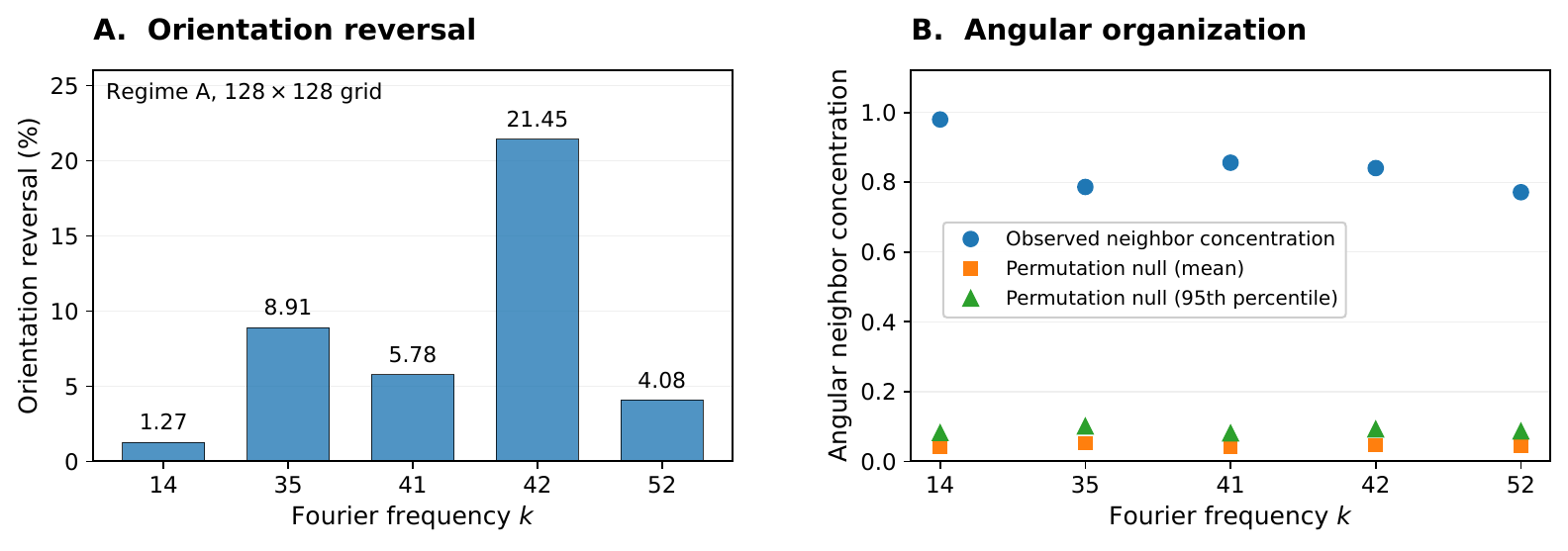}
\caption{Orientation reversal and angular organization across the five key Fourier frequencies. (A) Percentage of evaluation points with $|\mu|\geq 1$ on the $128\times128$ Regime A grid. (B) Observed concentration of the differences in $\arg\mu$ between neighboring grid points on the $16\times16$ Regime A grid, restricted to $|\mu|<1$, compared with the mean and 95th percentile of a spatial permutation null (1,000 permutations).}
\label{fig:B1}
\end{figure}

On the finer $128\times128$ Regime~A grid, the fractions with $|\mu|\geq1$ are $1.27\%$, $8.91\%$, $5.78\%$, $21.45\%$, and $4.08\%$ (Figure~\ref{fig:B1}A). For $k=14$, all 208 such points lie on three toroidal diagonals: 128 on $\theta_b-\theta_a=\pi$ (grid-index offset 64), and 40 on each adjacent diagonal (offsets 63 and 65). No reversing point lies outside this band. The percentages characterize this sampled resolution, not a resolution-independent bound.

\FloatBarrier

\subsection*{B.2. Measurements at real-embedding basepoints}
Regime~A evaluates the Fourier-component restriction on an artificial torus. Regime~B instead fixes the basepoint at the embedding of an actual modular-addition input pair while taking derivatives along the same Fourier-plane directions. Forty sampled pairs were evaluated at each frequency, and percentile-bootstrap intervals for the median were formed using 2,000 replicates (Table~\ref{tab:B2}).

\begin{table}[!htbp]
\centering
\caption{Directional Beltrami distortion around real input embeddings (Regime B). Each frequency is evaluated at 40 sampled input pairs; confidence intervals are percentile-bootstrap intervals for the median from 2,000 replicates. The derivative directions remain interventional.}
\label{tab:B2}
\vspace{0.25em}
\small
\begin{tabular}{@{}r r c r@{}}
\toprule
$k$ & Median $|\mu^{(a)}|$ & 95\% bootstrap CI & Fraction $|\mu|<1$ \\
\midrule
14 & 0.3223 & $[0.3035,\,0.3481]$ & 1.000\\
35 & 0.4432 & $[0.3112,\,0.6419]$ & 0.875\\
41 & 0.3077 & $[0.2942,\,0.3212]$ & 1.000\\
42 & 0.4525 & $[0.3632,\,0.8923]$ & 0.675\\
52 & 0.3550 & $[0.2642,\,0.6046]$ & 0.975\\
\bottomrule
\end{tabular}
\end{table}

The Regime~B medians ($0.3077$--$0.4525$) are similar in magnitude to the Regime~A estimates, although the orientation-preserving fractions vary between $0.675$ and $1.000$. This agreement shows that the measured magnitude does not require \emph{the basepoints themselves} to be outside the model's input distribution. It does not make the derivative measurements in-distribution: the infinitesimal Fourier-plane perturbations remain interventional. The 40-pair bootstrap characterizes uncertainty in the sampled median, not extreme values or global regularity.

\subsection*{B.3. Spatial organization of the Beltrami direction}
The argument of the complex coefficient, $\arg\mu$, describes the direction of local anisotropy; because the associated geometric axis is axial, a small global angular resultant need not imply spatially unstructured distortion. On the $16\times16$ Regime~A grid, the global axial resultant lengths range from $0.0124$ to $0.2019$ (Table~\ref{tab:B3}). To distinguish spatial arrangement from the marginal distribution of angles, we evaluated the concentration of angular differences between toroidally neighboring points, considering both grid directions and restricting the calculation to $|\mu|<1$.

\begin{table}[!htbp]
\centering
\caption{Angular organization of the directional Beltrami field in Regime A (epoch 40,000). The global axial resultant summarizes angular concentration across the grid, whereas neighboring concentration measures spatial organization of differences in $\arg\mu$ on the $16\times16$ grid, restricted to $|\mu|<1$, with toroidal neighbors in both directions. The spatial-permutation null uses 1,000 permutations.}
\label{tab:B3}
\vspace{0.35em}
\small
\begin{tabular}{@{}r r r r r@{}}
\toprule
$k$ & \shortstack{Global axial\\resultant $\overline R$} & \shortstack{Observed neighboring\\concentration} & \shortstack{Permutation\\null mean} & \shortstack{Permutation\\null 95th percentile} \\
\midrule
14 & 0.0134 & 0.9800 & 0.0412 & 0.0815 \\
35 & 0.2019 & 0.7866 & 0.0524 & 0.1003 \\
41 & 0.0124 & 0.8561 & 0.0420 & 0.0806 \\
42 & 0.0722 & 0.8407 & 0.0470 & 0.0919 \\
52 & 0.1592 & 0.7715 & 0.0447 & 0.0859 \\
\bottomrule
\end{tabular}
\par\vspace{0.42em}
\begin{minipage}{0.97\linewidth}\footnotesize\emph{Note.} At every frequency, the observed neighboring concentration exceeded all 1,000 reported spatial-permutation values. The null reassigns angular values to grid positions while retaining their marginal distribution.\end{minipage}
\end{table}

The observed neighboring concentrations are $0.7715$--$0.9800$, whereas the permutation-null means are $0.0412$--$0.0524$ (Figure~\ref{fig:B1}B; Table~\ref{tab:B3}). At each frequency, the observed statistic exceeds all 1,000 spatial permutations. Because the null preserves the collection of angular values while reassigning their positions, this result supports \emph{local spatial organization}, not global smoothness or concentration around one preferred axis. We also tested whether the low global resultants were forced by a putative identity $G_k(\lambda z,\overline\lambda w)=G_k(z,w)$ for $|\lambda|=1$. The median relative error on the Regime~A grid was $0.77$--$1.38$ across the five frequencies; the proposed identity does not hold in those measurements. The low global resultants therefore should not be attributed to that exact equivariance relation.
\FloatBarrier

\subsection*{B.4. Additional consistency checks}
Three further checks delimit the interpretation. First, exchanging the two input slots gives median differences $|\mu^{(a)}(z,w)-\mu^{(b)}(w,z)|$ between $0.0001$ and $0.0018$ across frequencies on the Regime~A grid. The map $G_k$ itself was tested separately and was approximately, but not exactly, symmetric; closeness of two normalized derivatives does not imply closeness of the underlying functions. Second, moving from $12\times12$ to $16\times16$ changed the per-frequency median $|\mu|$ by at most $0.0216$; it did not establish corresponding stability for maxima, upper quantiles, or orientation fractions. Third, in a basepoint-perturbation check limited to $k=14$ and $k=41$, perturbations up to $10^{-2}$ changed $|\mu|$ by $0.0000$ at the stated precision, while perturbations of $10^{-1}$ changed $|\mu|$ by at most $0.028$. Those perturbation results cannot be generalized to the other three frequencies.
\FloatBarrier

\subsection*{B.5. ReLU activation regions and local coefficient jumps}
The fine-grid maps in Figure~2 of the main text contain sharp features. Because the transformer's feedforward block is piecewise linear in its activations, we compared the complex-coefficient jumps across neighboring toroidal evaluation points with the associated changes in the ReLU activation pattern. For each of $k=14$ and $k=42$, the check considers 32,768 directed neighbor pairs on the $128\times128$ grid, one pair from every grid point along each of the two toroidal directions. It records $|\mu(x)-\mu(y)|$ and whether the activation patterns agree at the two endpoints.

\begin{table}[!htbp]
\centering
\caption{ReLU activation-pattern changes and local jumps of the complex directional Beltrami coefficient in Regime A (epoch 40,000; $128\times128$ grid). For each $k$, 32,768 directed toroidal neighboring pairs (one per grid point in each of two directions) are compared. Jumps are $|\mu(x)-\mu(y)|$; ``same'' and ``changed'' refer to equality of the feedforward ReLU activation patterns at the two points.}
\label{tab:B4}
\vspace{0.35em}
\begingroup
\setlength{\tabcolsep}{4.35pt}
\footnotesize
\begin{tabular}{@{}r r r r r r r r@{}}
\toprule
$k$ & \shortstack{Same\\pattern (\%)} & \shortstack{Median jump\\same} & \shortstack{Median jump\\changed} & \shortstack{99th pct.\\same} & \shortstack{99th pct.\\changed} & \shortstack{Top 1\% jumps\\with change} & \shortstack{Spearman\\$\rho$} \\
\midrule
14 & 62.8 & 0.0000 & 0.0558 & 0.0001 & 1.8951 & 1.000 & 0.731 \\
42 & 22.8 & 0.0261 & 0.0355 & 0.1354 & 0.3589 & 1.000 & 0.275 \\
\bottomrule
\end{tabular}
\endgroup
\par\vspace{0.45em}
\begin{minipage}{0.97\linewidth}\footnotesize\emph{Note.} Spearman $\rho$ is the correlation between the Hamming distance of activation patterns and the magnitude of the complex-coefficient jump; the ``top 1\%'' column is the fraction of the largest jumps with a changed pattern. These statistics are reported at the precision of the original ReLU check. They establish association, not a causal attribution of all jumps to ReLU boundaries.\end{minipage}
\end{table}

At $k=14$, the median jump is $0.0000$ at the reported precision for unchanged patterns versus $0.0558$ for changed patterns, with Spearman $\rho=0.731$ between activation-pattern Hamming distance and jump magnitude. The association is weaker at $k=42$ ($\rho=0.275$), where the median jumps are $0.0261$ and $0.0355$, respectively. At both frequencies, all of the largest 1\% of jumps connect points with different activation patterns (Table~\ref{tab:B4}). These observations associate sharp spatial changes with activation-pattern transitions; they neither identify a unique causal mechanism for each jump nor show that orientation reversal itself is caused by a ReLU boundary.

\endgroup
\clearpage
\newgeometry{margin=0.82in,top=0.70in,bottom=0.69in}
\begingroup
\linespread{1.0}\fontsize{10}{12}\selectfont
\renewcommand{\small}{\fontsize{9}{11}\selectfont}
\renewcommand{\footnotesize}{\fontsize{8}{9.5}\selectfont}
\setlength{\parskip}{4.3pt}
\setcounter{figure}{0}
\setcounter{table}{0}
\renewcommand{\thefigure}{C\arabic{figure}}
\renewcommand{\thetable}{C\arabic{table}}
\section*{Appendix C. Control Analyses}\label{app:C}
Section~3.2 compares the key Fourier frequencies with randomly constructed input planes and with a sixth, non-key Fourier frequency. This appendix specifies the plane comparison, documents its limitations, and presents the scale-sensitivity analysis. These are comparisons of directional Beltrami measurements \emph{within the same trained model}, not tests against an independently sampled population of networks.

\subsection*{C.1. Construction and aggregation of the control planes}
We compared the five key Fourier frequencies $k\in\{14,35,41,42,52\}$ with twenty random planes at the epoch-40,000 checkpoint. Each random input plane was formed from an orthonormal pair of functions on $\mathbb{Z}/113\mathbb{Z}$, as in the Fourier-plane construction, and rescaled to match the median input-frame norm of the key planes, $s_\star=0.51423$. The reference comparison was evaluated on a $12\times12$ grid. Norm matching is important because directional distortion can depend on the scale at which the input directions are probed. However, matching input norms does not make the input or readout frames equally conditioned.

The summaries in Table~\ref{tab:C1} require attention to their aggregation operators. The key-plane $|\mu|$ entry gives the \emph{range} of five frequency-level medians, whereas the random-plane entry is the \emph{median of twenty plane medians}. For orientation preservation, the reported key-plane figure is a mean over five frequency-level fractions and the random-plane figure is the median of twenty fractions. The two orientation entries therefore must not be treated as like-for-like estimators of one population quantity.

\subsection*{C.2. Separation, conditioning, and energy neutrality}
The five key-frequency medians span $0.302$--$0.435$, compared with $0.697$ for the random-plane median of medians (Table~\ref{tab:C1}). The difference is distributional rather than absolute: three random controls have medians $0.401$, $0.430$, and $0.431$, overlapping the range of the five key frequencies. Their combined rank ordering gives a Mann--Whitney statistic $U=4$ ($p=4.517\times10^{-4}$), which we report descriptively. The key frequencies were selected for their established functional role, not sampled randomly; independence between the twenty control measurements in the same trained network has also not been established. Consequently, this rank contrast should not be read as a conventional population-level inference.

\begin{table}[!htbp]
\centering
\caption{Control-plane comparability and diagnostic summaries (epoch 40,000; $12\times12$ grid). Key frequencies are $k\in\{14,35,41,42,52\}$. The sixth frequency $k=31$ is shown separately; its orientation sign depends on a declared, externally unanchored convention.}\label{tab:C1}
\vspace{0.28em}
\begingroup\small\setlength{\tabcolsep}{4.2pt}
\begin{tabular}{@{}p{0.255\linewidth}p{0.235\linewidth}p{0.295\linewidth}p{0.13\linewidth}@{}}
\toprule
Diagnostic & Key Fourier planes & Random control planes & $k=31$ \\
\midrule
Number of planes & 5 & 20 & 1 \\
Median $|\mu|$ & 0.302--0.435 (range) & 0.697 (median of medians) & 1.219 \\
Orientation preserved & 0.903 (mean) & 0.799 (median) & 0.000 \\
Input-frame $\kappa[C\ S]$ & 1.003--1.027 (range) & 1.3445 median; [1.048, 2.672] & Not reported \\
Readout-frame $\kappa[u\ v]$ & 1.01--1.03 (range) & 1.05--2.79 (range) & Not reported \\
Key-subspace energy overlap & Not applicable & 0.0947 mean; 0.0944 median; [0.054, 0.142] & Not reported \\
\bottomrule
\end{tabular}
\endgroup
\par\vspace{0.35em}
\begin{minipage}{\linewidth}\footnotesize
\emph{Notes.} Plane-wise summary operators differ: key-plane orientation preservation is the mean of five fractions, whereas the random-control value is the median of twenty fractions; these entries should not be directly compared as like-for-like estimates. The twenty controls are norm-matched in the input frame, but not conditioning-matched. The key-subspace energy fraction for random controls is close to the isotropic expectation $10/113=0.0885$. Three control-plane medians (0.401, 0.430, 0.431) fall in the key-frequency range, so there is no strict separation. The rank contrast ($U=4$, $p=4.517\times10^{-4}$) is descriptive, not a population-level inferential test: the key modes were selected functionally, and independence among controls within the same trained network has not been established. Orientation at $k=31$ is reported under a convention without a direct sign anchor.
\end{minipage}
\end{table}

The random controls have median input-frame condition number $\kappa[C\ S]=1.3445$ (range $1.048$--$2.672$), versus $1.003$--$1.027$ for the key planes. Their readout-frame condition numbers span $1.05$--$2.79$, versus $1.01$--$1.03$ for the keys. Thus, lower key-frequency distortion cannot be uniquely attributed to functional circuit membership on the basis of norm matching alone: differences in frame conditioning remain an alternative contributor. A separate energy check found that the random input-plane pairs project onto the ten-dimensional key-frequency subspace with mean energy fraction $0.0947$ (median $0.0944$, range $0.054$--$0.142$), close to the isotropic reference $10/113=0.0885$. This supports neutrality with respect to key-subspace energy, but does not remove the conditioning difference.
\FloatBarrier

\subsection*{C.3. Sensitivity to the input-plane scale}
Because the main comparison fixes one reference norm, we examined five scale multipliers $m\in\{0.50,0.75,1.00,1.50,2.00\}$ (Figure~\ref{fig:C1}). For random controls, the input frames were normalized to $m s_\star$; for each key frequency, its existing input frame was scaled by $m$. These are two distinct scaling constructions, and $m$ is therefore the appropriate axis for comparing the resulting sensitivity curves, not a claim of identical input frames.

Across the five multipliers, the random controls have median-of-medians $0.687$, $0.735$, $0.697$, $0.708$, and $0.712$, respectively. The key-frequency medians remain approximately $0.27$--$0.45$ across the tested scales. Individual overlap is retained at the three smaller multipliers: two to three random controls fall in the key-frequency range for $m\leq1.00$, whereas none do at $m=1.50$ or $2.00$. The random-control median remains above the key-frequency range at every tested multiplier. The conclusion is consequently \emph{robust in the tested sweep at the group-summary level}, not a universal separation between individual planes or scales outside that range.

\begin{figure}[!htbp]
\centering
\includegraphics[width=0.97\linewidth]{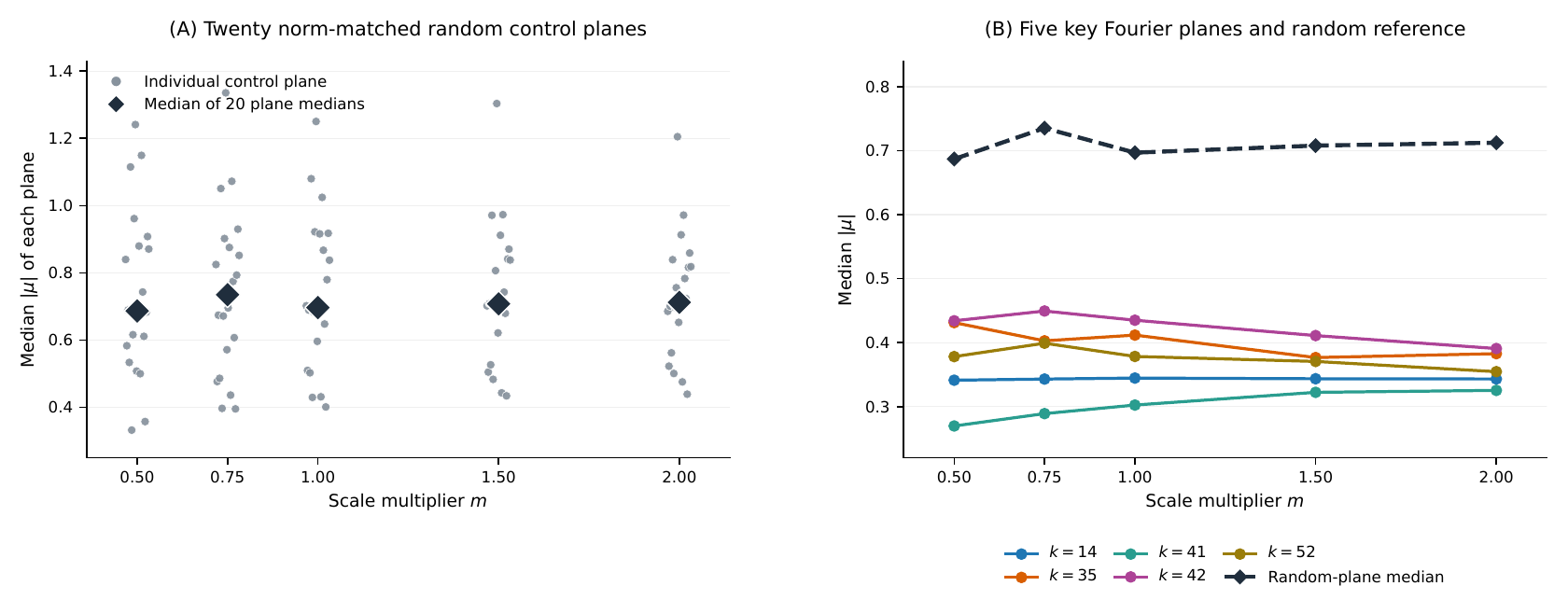}
\caption{Scale sensitivity of the key Fourier planes and the twenty norm-matched random controls (epoch 40,000; $12\times12$ grid). The multiplier $m$ ranges from $0.50$ to $2.00$: random control input frames are normalized to $m s_\star$ ($s_\star=0.51423$), while each key-frequency input frame is multiplied by $m$ relative to its original scale. (A) Each gray point is one random plane's median $|\mu|$; dark diamonds are the medians of the twenty per-plane medians. Horizontal jitter is only for visibility. (B) Median $|\mu|$ for each of the five fixed key-frequency planes across scale multipliers; the dashed curve repeats the random-plane median-of-medians for comparison. Lines connect measurements of the \emph{same} frequency at different scale multipliers. At $m\leq1.00$, two to three individual random planes fall within the key-frequency range; none do at $m=1.50$ or $2.00$. The control-group median remains higher at every tested multiplier.}\label{fig:C1}
\end{figure}
\FloatBarrier

\subsection*{C.4. The additional Fourier frequency $k=31$}
The additional frequency $k=31$ is present in the embedding matrix but has much weaker readout vectors than the five functionally identified key frequencies. In the $12\times12$ plane comparison, its median $|\mu|$ is $1.219$ and its sampled orientation-preserving fraction is zero (Table~\ref{tab:C1}). Its determinant sign is constant on the sampled Regime~A grid, unlike the mixtures of signs measured at the five key frequencies and the twenty random controls. The \emph{constancy} of the sign is a geometric observation; the designation of the constant sign as orientation-reversing is contingent on a readout-orientation convention that has no direct external anchor for $k=31$. Under the conjugate readout convention, every sampled determinant sign at $k=31$ flips and $|\mu|$ becomes $1/|\mu|$ (median approximately $0.820$ instead of $1.219$); the constancy of the sign is unchanged (Appendix~A.3). Neither the magnitude $|\mu|\approx1.22$ nor the chosen sign, considered alone, establishes a distinct algorithmic role for this frequency. We therefore report $k=31$ as a comparison case, not as another functional key frequency or a calibrated negative control.

\endgroup
\clearpage
\newgeometry{margin=0.82in,top=0.71in,bottom=0.73in}
\begingroup
\linespread{1.0}\fontsize{10}{12}\selectfont
\renewcommand{\small}{\fontsize{9}{11}\selectfont}
\renewcommand{\footnotesize}{\fontsize{8}{9.5}\selectfont}
\setlength{\parskip}{4.6pt}
\setcounter{figure}{0}
\setcounter{table}{0}
\renewcommand{\thefigure}{D\arabic{figure}}
\renewcommand{\thetable}{D\arabic{table}}
\section*{Appendix D. Replication Across Training Runs}\label{app:D}
Section~3.2 reports that the reduced directional Beltrami distortion of the key Fourier representations recurs after independent retraining. This appendix describes how frequencies were selected within each run, presents the paired measurements at epochs 40,000 and 40,100, and makes explicit the criteria behind the range of replication counts. Replication here concerns the \emph{distribution of frequency-level geometric measurements}, not the identity of a frequency index across independently initialized networks.

\subsection*{D.1. Independent retrainings and measurement protocol}
Six independent seeds were trained with the architecture and hyperparameters of the main experiment ($P=113$, model width 128, four attention heads, MLP width 512, training fraction 0.3, weight decay 1) to epoch 40,100. For each run, the five frequencies with the largest Fourier-component norms of the embedding matrix $W_E$ (columns 0--112) were selected using the same rule, and this selection was held fixed across the two measured checkpoints. The selected indices vary between seeds; equal values of $k$ in different runs should \emph{not} be interpreted as identifying the same learned functional object.

Measurements followed Regime~A on the $16\times16$ angular grid. For each selected frequency and epoch, we record the full-grid median $|\mu|$, the fraction $f_+$ of points with $|\mu|<1$, and the median dilatation $K=(1+|\mu|)/(1-|\mu|)$ restricted to that orientation-preserving locus (Table~\ref{tab:D1}). When this locus is empty, median $K$ is undefined rather than zero. The complete three-panel comparison is shown in Figure~\ref{fig:D1}.

\subsection*{D.2. Concentration and departures from the central range}
At epoch 40,000, 22 of the 30 frequency-level medians fall in $[0.30,0.45]$; one is below $0.30$, three are in $(0.45,0.90]$, and four exceed $0.90$ (Table~\ref{tab:D1}; Figure~\ref{fig:D1}, top). The central concentration recurs across retrainings, but it is not universal: for example, seed~6 at $k=43$ has median $|\mu|=1.7480$ and $1.9400$ at the two checkpoints and no orientation-preserving grid point at either epoch. For seed~1, $k=35$ changes from $0.9076$ to $1.1050$, so its median $K$ is undefined at epoch 40,100. These cases are retained in both the visualization and the complete table instead of being censored from the replication summary.

The two checkpoints are only 100 epochs apart, and Figure~\ref{fig:D1} depicts their paired variation rather than treating them as independent observations. Moreover, a frequency can meet a criterion at epoch 40,000 without meeting it at both epochs. This distinction is necessary when interpreting the band counts below.

\subsection*{D.3. Sensitivity to the replication criterion}
The 22/30 value in Section~3.2 is a \emph{single-epoch} count of medians in $[0.30,0.45]$ and does not restrict $K$. The bottom-right panel of Figure~\ref{fig:D1} instead applies each specified criterion \emph{at both epochs 40,000 and 40,100} to the same thirty frequency-level records. With medians and $K$ ranges inclusive of their endpoints, the five criteria are:
\begin{enumerate}\setlength{\itemsep}{1pt}\setlength{\topsep}{2pt}
\item \textbf{Strict:} $\operatorname{med}|\mu|\in[0.30,0.45]$ and $K\in[1.5,2.6]$: \textbf{20/30}.
\item \textbf{$|\mu|$ only:} $\operatorname{med}|\mu|\in[0.30,0.45]$ with $K$ unrestricted: \textbf{21/30}.
\item \textbf{$\pm0.05$:} $\operatorname{med}|\mu|\in[0.25,0.50]$ and $K\in[1.4,2.8]$: \textbf{23/30}.
\item \textbf{Tolerant:} $\operatorname{med}|\mu|\in[0.20,0.55]$ and $K\in[1.3,3.0]$: \textbf{25/30}.
\item \textbf{$|\mu|<1$ only:} $\operatorname{med}|\mu|\in[0,1]$ with $K$ unrestricted: \textbf{27/30}.
\end{enumerate}
These totals range from 20 to 27 of 30 because they answer different numerical questions; they do not contradict the 22/30 single-epoch observation. The six-run dataset does not by itself warrant a population-level significance claim, and the observed concentration is not evidence that lower distortion is unique to functional key modes relative to all alternative directions (see Appendix~C).

\clearpage
\begin{minipage}{\linewidth}
\centering
\includegraphics[width=0.92\linewidth]{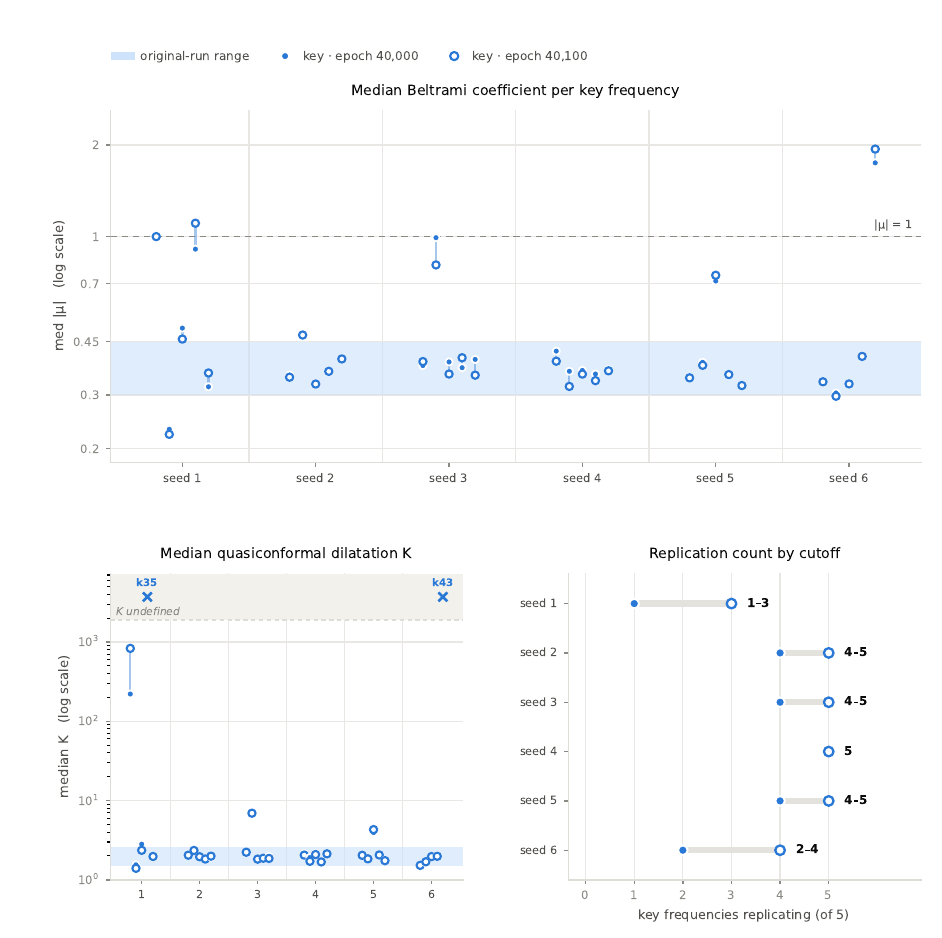}
\captionof{figure}{\textbf{Geometric measurements from six independently trained seeds.} Each run contributes its five highest-embedding-Fourier-norm frequencies, selected independently of the other seeds. Filled and hollow markers denote epochs 40,000 and 40,100; paired observations are joined. \emph{Top:} median $|\mu|$ on the Regime~A $16\times16$ grid; the blue band marks $[0.30,0.45]$ and the dashed line marks $|\mu|=1$. \emph{Bottom left:} median $K$ on $|\mu|<1$ (blue band: $1.5$--$2.6$); the gray strip identifies cases with undefined $K$. \emph{Bottom right:} counts per seed under five nested, two-epoch acceptance bands, from strict to permissive. Their exact definitions and aggregate totals (20--27 of 30) appear in Section~D.3.}\label{fig:D1}
\end{minipage}
\vspace{0.7em}
\subsection*{D.4. Data provenance and scope of the replication}
Four runs were trained and measured in one computational environment. Two additional runs were contributed with their measurement files but without the checkpoints; their SHA-256 checksums were reported but could not be recomputed from the received materials. This limits independent checkpoint-level verification for those two seeds, without changing the tabulated measurements. Training accuracy reached 100\% for all runs; the separate full-pair forward-pass accuracies at the two reported epochs appear in the notes to Table~\ref{tab:D1}. The experiment supports recurrence of the distributional pattern \emph{across these six retrainings}, not a claim of identical geometry at every key frequency or of inferential certainty from six seeds.

\clearpage
\begin{table}[!htbp]
\centering
\caption{\textbf{Complete per-seed directional Beltrami measurements.} Five independently selected frequencies per seed, evaluated at epochs 40,000 and 40,100 (Regime~A, $16\times16$ grid). The $K$ column is the median on $|\mu|<1$; $f_+$ is the fraction of grid points in that locus.}\label{tab:D1}
\vspace{0.30em}
\begingroup\footnotesize\setlength{\tabcolsep}{5.1pt}
\begin{tabular*}{0.97\textwidth}{@{\extracolsep{\fill}}cc rrr rrr@{}}
\toprule
& & \multicolumn{3}{c}{Epoch 40,000} & \multicolumn{3}{c}{Epoch 40,100}\\
\cmidrule(lr){3-5}\cmidrule(lr){6-8}
Seed & $k$ & Med. $|\mu|$ & Med. $K$ & $f_+$ & Med. $|\mu|$ & Med. $K$ & $f_+$\\
\midrule
1 & 3 & 1.0010 & 221 & 0.4961 & 0.9982 & 830 & 0.7539 \\
 & 6 & 0.2312 & 1.546 & 0.8828 & 0.2224 & 1.397 & 0.8711 \\
 & 23 & 0.4983 & 2.816 & 0.8164 & 0.4584 & 2.356 & 0.7305 \\
 & 35 & 0.9076 & 20.472 & 0.9688 & 1.1050 & \textemdash & 0.0000 \\
 & 39 & 0.3192 & 1.936 & 0.9883 & 0.3544 & 1.969 & 0.9492 \\
\addlinespace[0.25em]\midrule
2 & 6 & 0.3488 & 2.069 & 0.9844 & 0.3433 & 2.044 & 0.9883 \\
 & 7 & 0.4666 & 2.366 & 0.8125 & 0.4728 & 2.345 & 0.8086 \\
 & 15 & 0.3251 & 1.954 & 0.9922 & 0.3258 & 1.954 & 0.9922 \\
 & 27 & 0.3570 & 1.829 & 0.8086 & 0.3588 & 1.828 & 0.8086 \\
 & 38 & 0.3943 & 1.939 & 0.8477 & 0.3944 & 1.988 & 0.8516 \\
\addlinespace[0.25em]\midrule
3 & 9 & 0.3750 & 2.199 & 0.9961 & 0.3865 & 2.228 & 0.9805 \\
 & 18 & 0.9901 & 6.869 & 0.5078 & 0.8051 & 6.928 & 0.6328 \\
 & 27 & 0.3854 & 1.830 & 0.8203 & 0.3515 & 1.824 & 0.8555 \\
 & 53 & 0.3690 & 1.871 & 0.8398 & 0.3979 & 1.872 & 0.8516 \\
 & 55 & 0.3929 & 1.966 & 0.8984 & 0.3486 & 1.862 & 0.8711 \\
\addlinespace[0.25em]\midrule
4 & 11 & 0.4184 & 2.120 & 0.8984 & 0.3876 & 2.050 & 0.8945 \\
 & 25 & 0.3590 & 1.910 & 0.8281 & 0.3200 & 1.713 & 0.8516 \\
 & 32 & 0.3614 & 2.132 & 1.0000 & 0.3515 & 2.084 & 1.0000 \\
 & 40 & 0.3517 & 1.667 & 0.8398 & 0.3344 & 1.683 & 0.8594 \\
 & 49 & 0.3635 & 2.142 & 1.0000 & 0.3603 & 2.126 & 1.0000 \\
\addlinespace[0.25em]\midrule
5 & 9 & 0.3419 & 2.039 & 0.9961 & 0.3416 & 2.037 & 0.9961 \\
 & 34 & 0.3844 & 1.834 & 0.7891 & 0.3758 & 1.837 & 0.7891 \\
 & 37 & 0.7122 & 3.943 & 0.7227 & 0.7442 & 4.300 & 0.7266 \\
 & 38 & 0.3517 & 2.032 & 0.9766 & 0.3502 & 2.057 & 0.9844 \\
 & 55 & 0.3222 & 1.760 & 0.8047 & 0.3222 & 1.745 & 0.7969 \\
\addlinespace[0.25em]\midrule
6 & 7 & 0.3300 & 1.492 & 0.7812 & 0.3314 & 1.521 & 0.7852 \\
 & 11 & 0.3045 & 1.645 & 0.8477 & 0.2971 & 1.689 & 0.8477 \\
 & 35 & 0.3246 & 1.961 & 1.0000 & 0.3261 & 1.963 & 0.9727 \\
 & 41 & 0.4014 & 2.028 & 0.8203 & 0.4020 & 1.982 & 0.8125 \\
 & 43 & 1.7480 & \textemdash & 0.0000 & 1.9400 & \textemdash & 0.0000 \\
\bottomrule
\end{tabular*}
\endgroup
\vspace{0.38em}
\begin{minipage}{0.98\linewidth}\footnotesize
\emph{Notes.} A dash denotes undefined median $K$ because no grid point has $|\mu|<1$. Fractions $f_+$ are computed from 256 grid points. The five frequency indices were selected separately for each training seed and are not comparable across seeds. Full-pair forward-pass accuracy (epochs 40,000 / 40,100): seed~1, 99.4675\% / 99.7416\%; seed~2, 99.9843\% / 99.9843\%; seed~3, 99.9530\% / 99.9687\%; seeds~4--6, 100\% / 100\%. Four runs were measured in one environment, and two were contributed externally without their checkpoints. The latter checksums were reported but not independently recomputed.
\end{minipage}
\end{table}

\endgroup
\clearpage
\newgeometry{margin=0.82in,top=0.58in,bottom=0.50in}
\begingroup
\linespread{0.98}\fontsize{10}{12}\selectfont
\renewcommand{\small}{\fontsize{9}{11}\selectfont}
\renewcommand{\footnotesize}{\fontsize{8}{9.5}\selectfont}
\setlength{\parskip}{1.8pt}
\setcounter{figure}{0}
\setcounter{table}{0}
\setcounter{equation}{0}
\renewcommand{\theequation}{\arabic{equation}}
\renewcommand{\thefigure}{E\arabic{figure}}
\renewcommand{\thetable}{E\arabic{table}}
\renewcommand{\arraystretch}{1.06}
\section*{Appendix E. Training Dynamics and Behavioral Validation}\label{app:E}
Section~3.3 reports two tests that delimit what directional Beltrami distortion can reveal about the trained transformer: its evolution relative to Fourier-circuit formation and its association with behavior on individual input pairs. This appendix specifies the checkpoint and comparison rules, presents the complete quantitative summaries, and separates evidence of temporal ordering or spatial organization from evidence of a predictive or causal link to the learned computation.

\subsection*{E.1. Checkpoints, geometric change, and comparison windows}
We measured the median directional coefficient $M_k(t)=\operatorname{med}|\mu_k|$ on the Regime~A $16\times16$ angular grid for each of the five key frequencies, using all 500 saved checkpoints from epoch 0 to 49,900 in steps of 100. The normalization in the temporal comparison uses epoch 40,000, the trained checkpoint evaluated in the main results, rather than the final saved epoch. For a checkpoint $t$, the signed fraction of initial-to-final geometric change is
\begin{equation}
F_k(t)=\frac{M_k(0)-M_k(t)}{M_k(0)-M_k(40{,}000)}.
\label{eq:E1}
\end{equation}
The ratio is not clipped to $[0,1]$: it can be negative if distortion initially increases, or exceed one if it passes its epoch-40,000 level. Its interpretation depends on a non-negligible denominator.

The reference epochs were determined from the loss curves before the Beltrami trajectories were evaluated. Memorization was defined by the first training-loss crossing below $10^{-3}$ (epoch 476); the onset and end of the test-loss generalization transition (epochs 2,390 and 9,830) were defined by a 5\% and 95\% decline, respectively, in the test-loss range measured from its post-memorization maximum at epoch 1,420 (test loss recorded every 10 epochs). Because checkpoints were saved every 100 epochs, epoch 2,300 is the last saved checkpoint strictly preceding the primary onset at 2,390. The declared geometric criterion required $F_k\geq0.60$ for at least four of the five key frequencies before that boundary. A second, earlier cutoff at epoch 500 was fixed as a sensitivity check. Epoch 2,400, which is just after the primary boundary, is reported separately to match the comparison in Section~3.3 and Figure~4.

\paragraph{Checkpoint and script provenance.}
The source model is the publicly released \texttt{full\_run\_data.pth} archive (456,428,905 bytes; SHA-256 \texttt{\seqsplit{a151f02125a33d6647d7b27fce99ae145ff3b83abc0586a883acc73e8318d828}}). Its epoch-40,000 state was verified against all $113^2=12,769$ ordered input pairs, yielding 100\% forward-pass accuracy and a mean cross-entropy of $2.4122\times10^{-7}$, as documented in the original experimental verification.

The original scripts for the circuit-variance and behavioral-margin analyses were not retained. The analysis scripts \texttt{circuito.py} and \texttt{puente.py} are reconstructions, not the original implementations; their recorded results reproduce the circuit fractions (0.0277 at epoch 2,400 and 0.9690 at epoch 40,000), the five reported Spearman coefficients, and test accuracy 0.509789. The archived numerical records used to construct Tables~\ref{tab:E1} and~\ref{tab:E2} are \texttt{serie500.json}, \texttt{circuito500.json}, and \texttt{puente.json}; no interpolation was applied. The maximum $|\rho|=0.052$ over additional configurations is documented in the experimental methods note rather than the primary \texttt{puente.json} summary.

\subsection*{E.2. Directional distortion before Fourier-circuit formation}
The Fourier-circuit metric measures the fraction of pair-centered logit variance explained by the five key frequencies. For each of the $113^2$ ordered input pairs $(a,b)$, the mean over the 113 output logits is first subtracted. Each resulting output-logit series is then centered across all input pairs and projected onto the ten orthonormalized directions $\cos(\omega_k(a+b))$ and $\sin(\omega_k(a+b))$, for $k\in\{14,35,41,42,52\}$ and $\omega_k=2\pi k/113$. The metric is the sum of squared norms of these projections, over output logits, divided by the sum of squared norms of the corresponding centered logit series.

Table~\ref{tab:E1} reports the frequency-level measurements and the Fourier-circuit metric at matched checkpoints. At epoch 2,300, the signed fractions exceed $0.60$ for all five frequencies under the declared numerical rule. At epoch 500, four of five exceed the threshold: $k=14$ has $F_{14}(500)=-0.102$, indicating a small increase rather than an accumulated reduction at that earlier checkpoint. For the four frequencies with a non-degenerate initial-to-final difference, the fractions at epoch 2,400 range from $0.745$ to $0.966$. They therefore substantially precede the main rise in the measured circuit metric.

\begin{table}[!htbp]
\centering
\caption{\textbf{Geometric change and Fourier-circuit formation at reference checkpoints.} \emph{(A)} $M_k(t)=\operatorname{med}|\mu_k|$ on the Regime~A $16\times16$ grid, and signed fractions $F_k(t)$ defined in Eq.~\ref{eq:E1}. Epoch 2,300 is the last saved checkpoint strictly before generalization onset (epoch 2,390); epoch 500 is the sensitivity cutoff, and epoch 2,400 is the nearest saved checkpoint afterward. \emph{(B)} Fraction of pair-centered logit variance explained by the five key frequencies at the same checkpoints.}
\label{tab:E1}
\vspace{0.13em}
\textbf{(A) Signed fraction of geometric change}\par\vspace{0.13em}
\begingroup\small\setlength{\tabcolsep}{7.3pt}
\begin{tabular}{@{}rccccc@{}}
\toprule
$k$ & $M_k(0)$ & $M_k(40{,}000)$ & $F_k(500)$ & $F_k(2{,}300)$ & $F_k(2{,}400)$\\
\midrule
14 & 0.4758 & 0.3468 & $-0.102$ & 0.746 & 0.745 \\
35 & 1.1289 & 0.3902 & 0.859 & 0.963 & 0.966 \\
41 & 2.2779 & 0.3036 & 0.957 & 0.950 & 0.948 \\
42 & 0.9394 & 0.4455 & 0.693 & 0.777 & 0.773 \\
52 & 0.4035 & 0.4001 & $47.8^{\dagger}$ & $43.9^{\dagger}$ & $42.4^{\dagger}$ \\
\bottomrule
\end{tabular}
\endgroup
\par\vspace{0.24em}
\textbf{(B) Fourier-circuit explained logit variance}\par\vspace{0.13em}
\begingroup\small\setlength{\tabcolsep}{17pt}
\begin{tabular}{@{}rc@{}}
\toprule
Epoch & Explained logit variance\\
\midrule
0 & $4.5058\times10^{-5}$ \\
500 & 0.00963 \\
2,300 & 0.02669 \\
2,400 & 0.02775 \\
40,000 & 0.96902 \\
\bottomrule
\end{tabular}
\endgroup
\par\vspace{0.2em}
\begin{minipage}{0.98\linewidth}\footnotesize
\emph{Notes.} Under the primary cutoff, $F_k(2{,}300)\geq0.60$ for 5/5 frequencies (4/4 after excluding $k=52$); at epoch 500 the count is 4/5 (3/4 without $k=52$). $^\dagger$For $k=52$, $M_k(0)-M_k(40{,}000)=0.00345951$, small relative to its observed training variation, so the reported ratios are numerically defined but not informative effect sizes. No fractions are truncated to $[0,1]$.
\end{minipage}
\end{table}
\FloatBarrier

At epoch 2,400, the Fourier-circuit metric is $0.02775$, compared with $0.96902$ at epoch 40,000, or approximately $2.9\%$ of the latter. Its much later increase is also apparent in Figure~4B; by contrast, the individual Beltrami trajectories are not uniformly monotonic during training (Figure~4A). The numerical criterion has an important limitation: the initial-to-final difference for $k=52$ is only $0.00345951$, while the frequency's median varies considerably across the training trajectory. Its very large fractions therefore pass the threshold mechanically rather than providing interpretable evidence of early change. Excluding $k=52$ preserves the primary count (4/4) but reduces the sensitivity count to 3/4. These comparisons document temporal ordering in this run; they neither identify what causes the geometric change nor show that the measured coefficient is used by the Fourier circuit.
\FloatBarrier

\subsection*{E.3. Pairwise association with behavioral margins}\enlargethispage{2\baselineskip}
The second test asks whether directional distortion predicts the correct-logit margin for an individual input pair. We chose the checkpoint by the previously fixed rule of test accuracy closest to $50\%$, without using the Beltrami values to select it. This gives epoch 9,400, with test accuracy $0.509789$ and training accuracy $1.0000$. The analysis uses all 8,939 held-out pairs, with the actual input pair as the base point for the directional evaluation, and no excluded observations. The association measure is Spearman's rank correlation between $|\mu_k|$ and the correct-logit margin; the predeclared threshold for a substantively meaningful effect was $|\rho|\geq0.20$.

\begin{table}[!htbp]
\centering
\caption{\textbf{Association of directional Beltrami magnitude with correct-logit margin.} Spearman correlations for the five key frequencies at epoch 9,400 (8,939 held-out pairs; no exclusions). The null threshold is the frequency-specific 95th percentile of $|\rho|$ over 1,000 label permutations; the permutation $p$-values are two-sided and unadjusted across frequencies.}
\label{tab:E2}
\vspace{0.13em}
\begingroup\small\setlength{\tabcolsep}{10.5pt}
\begin{tabular}{@{}rcccc@{}}
\toprule
$k$ & Spearman $\rho$ & Null 95th $|\rho|$ & Permutation $p$ & Exceeds null?\\
\midrule
14 & $-0.0260$ & 0.0206 & 0.010 & Yes\\
35 & $-0.0146$ & 0.0208 & 0.166 & No\\
41 & $-0.0070$ & 0.0212 & 0.507 & No\\
42 & $-0.0224$ & 0.0196 & 0.026 & Yes\\
52 & $-0.0058$ & 0.0208 & 0.582 & No\\
\bottomrule
\end{tabular}
\endgroup
\par\vspace{0.2em}
\begin{minipage}{0.98\linewidth}\footnotesize
\emph{Notes.} Two frequencies ($k=14$ and $k=42$) exceed their individual permutation-null 95th percentiles; no family-wise significance claim is made. None approaches the declared $|\rho|\geq0.20$ effect-size threshold. The non-geometric embedding-norm baseline, $\|W_E[:,a]\|+\|W_E[:,b]\|$, yields $\rho=-0.08946$ against the same margin.
\end{minipage}
\end{table}

The maximum frequency-level correlation is $|\rho|=0.0260$. The two unadjusted, frequency-specific null crossings ($k=14$ and $k=42$) are far below the declared effect-size threshold. The non-geometric embedding-norm baseline ($\rho=-0.08946$) is also small, but exceeds the magnitude of every Beltrami correlation. Three sensitivity analyses---epoch-9,900 test pairs, epoch-9,400 training pairs, and a full-logit variant---accompany the primary epoch-9,400 test-set analysis. Across these four configurations, the reported maximum is $|\rho|=0.052$. Only that aggregate maximum, not the additional per-frequency estimates, is documented. Under the tested conditions, $|\mu|$ is not a useful predictor of pairwise correct-logit margins.
\FloatBarrier

\FloatBarrier

\endgroup
\end{document}